\documentclass{article}

\usepackage{microtype}
\usepackage{graphicx}
\usepackage{subfigure}
\usepackage{booktabs} 

\usepackage[T1]{fontenc}
\usepackage{wrapfig}
\usepackage{amsmath, amsthm, amssymb, commath,dsfont}
\usepackage{shortbold}

\newtheorem{proposition}{Proposition}
\newtheorem{definition}{Definition}

\usepackage{hyperref}

\usepackage[accepted]{icml2019}

\icmltitlerunning{The Anisotropic Noise in Stochastic Gradient Descent}

\begin{document}

\twocolumn[
\icmltitle{The Anisotropic Noise in Stochastic Gradient Descent: Its Behavior of Escaping from Sharp Minima and Regularization Effects}



\icmlsetsymbol{equal}{*}

\begin{icmlauthorlist}
\icmlauthor{Zhanxing Zhu}{equal,sms,cds,bibdr}
\icmlauthor{Jingfeng Wu}{equal,sms}
\icmlauthor{Bing Yu}{sms}
\icmlauthor{Lei Wu}{sms}
\icmlauthor{Jinwen Ma}{sms}
\end{icmlauthorlist}

\icmlaffiliation{sms}{School of Mathematical Sciences, Peking University, Beijing, China}
\icmlaffiliation{cds}{Center for Data Science, Peking University, Beijing, China}
\icmlaffiliation{bibdr}{Beijing Institute of Big Data Research, Beijing, China}

\icmlcorrespondingauthor{Zhanxing Zhu}{zhanxing.zhu@pku.edu.cn}
\icmlcorrespondingauthor{Jingfeng Wu}{pkuwjf@pku.edu.cn}

\icmlkeywords{Machine Learning, ICML}

\vskip 0.3in
]



\printAffiliationsAndNotice{\icmlEqualContribution} 

\begin{abstract}
Understanding the behavior of stochastic gradient descent (SGD) in the context of deep neural networks has raised lots of concerns recently. 
Along this line, we  study a general form of gradient based optimization dynamics with unbiased noise, which unifies SGD and standard Langevin dynamics.
Through investigating this general optimization dynamics, we analyze the behavior of SGD on escaping from minima and its regularization effects. A novel indicator is derived to characterize the efficiency of escaping from minima through measuring the alignment of noise covariance and the curvature of loss function.
Based on this indicator, two conditions are established to show which type of noise structure is superior to isotropic noise in term of escaping efficiency.
We further show that the anisotropic noise in SGD satisfies the two conditions, and thus helps to  escape from sharp and poor minima effectively, towards more stable and flat minima that typically generalize well.
We systematically design various experiments to verify the benefits of the anisotropic noise, compared with full gradient descent plus isotropic diffusion (i.e. Langevin dynamics).
\end{abstract}

\section{Introduction}
\label{sec:intro}

As a successful learning algorithm, stochastic gradient descent (SGD) was originally adopted for dealing with the computational bottleneck of training neural networks with large-scale datasets~\citep{bottou1991stochastic}.
Its empirical efficiency and effectiveness have attracted lots of attention.
Besides the aspect of empirical efficiency, recently, researchers started to analyze the optimization behaviors of SGD and its impacts on generalization. 

\begin{figure}
\centering
\begin{tabular}{cc}
\includegraphics[width=0.45\linewidth]{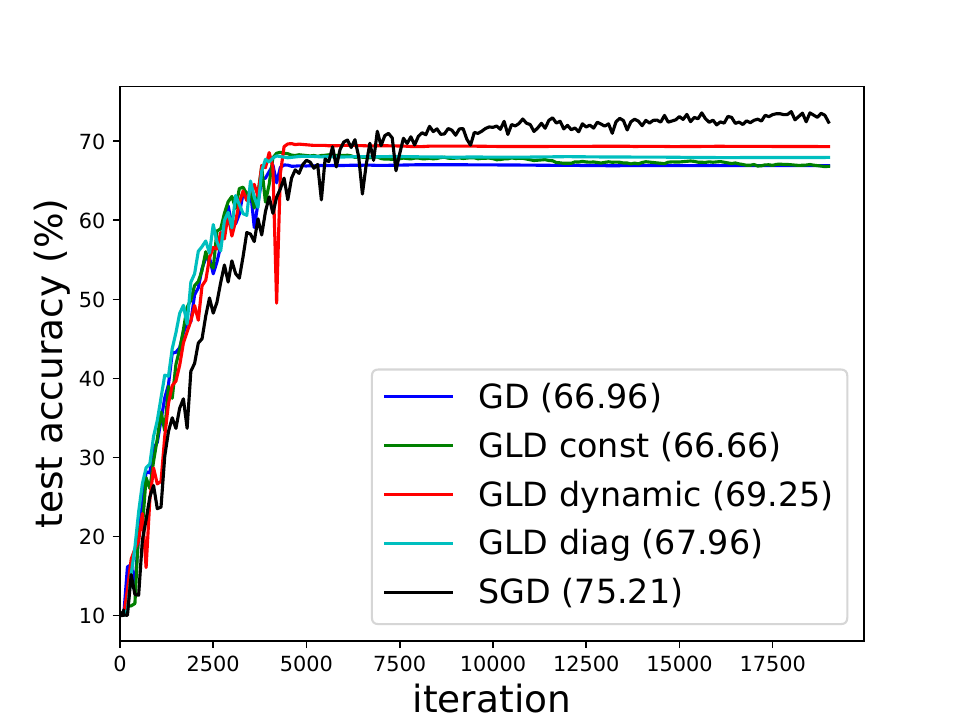} & \includegraphics[width=0.45\linewidth]{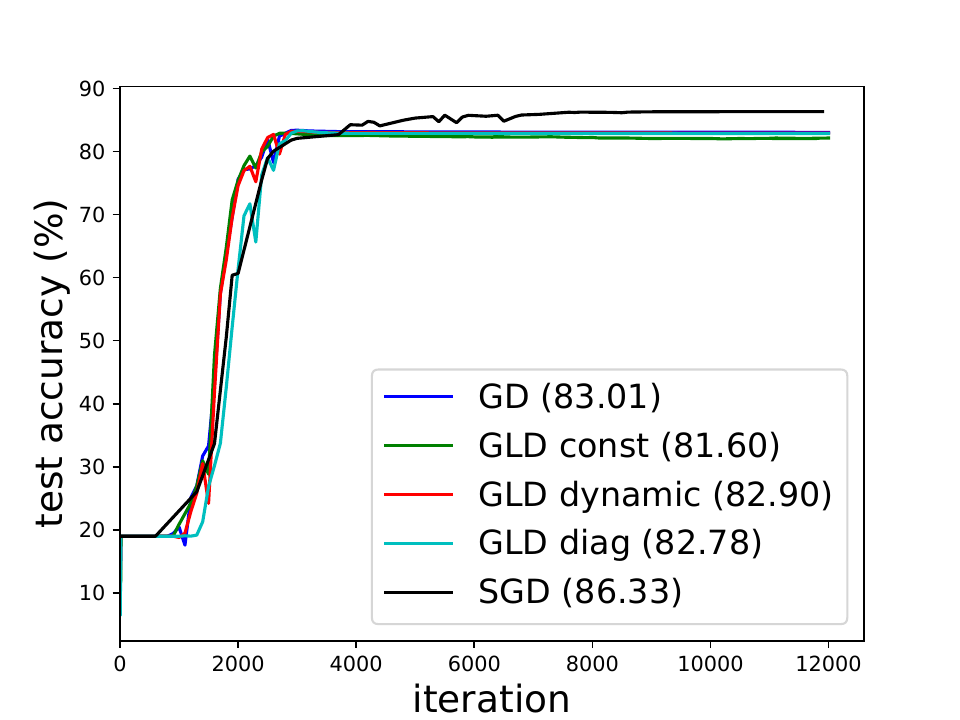}
\end{tabular}
\vspace{-4mm}
\caption{\small The generalization performance of dynamics in Table~\ref{tb:dynamics}.
The noise magnitude of SGD, GLD dynamic and GLD diag is tuned to be the same for fair comparison. The noise of GLD constant is tunded to the best.
\textbf{Left}: SVHN. We only use $2,5000$ examples for training to compromise with the computational burden;
\textbf{Right}: CIFAR-10.
The model is VGG-11 since it achieves decent performance without using batch normalization, which causes uncontrollable affects for analyzing SGD.
}
\label{fig:generalization_svhn_cifar}
\vspace{-6mm}
\end{figure}

The optimization properties of SGD have been studied from various perspectives.
The convergence behaviors of SGD for simple one hidden layer neural networks were investigated in~\citep{li2017convergence,brutzkus2017sgd}.
In non-convex settings, the characterization of how SGD escapes from stationary points, including  saddle points and local minima,  was analyzed in~\citep{daneshmand2018escaping,jin2017escape,hu2017diffusion}. 
On the other hand, in the context of deep learning, researchers realized that the noise introduced by SGD impacts the generalization, thanks to the research  on the phenomenon that training with a large batch could cause a significant drop of test accuracy~\citep{keskar2016large}.
Particularly, several works attempted to investigate how the magnitude of the noise influences the generalization during the process of SGD optimization, including the batch size and learning rate~\citep{hoffer2017,goyal2017accurate,chaudhari2017stochastic,jastrzkebski2017three}.  
Another line of research interpreted SGD from a Bayesian perspective.
In~\citep{mandt2017stochastic,chaudhari2017stochastic}, SGD was interpreted as performing variational inference, where certain entropic regularization involves to prevent overfitting.
And the work~\citep{smith2018bayesian} attempted to provide an understanding based on model evidence. These explanations are compatible with the flat/sharp minima argument~\citep{hochreiter1997flat,keskar2016large}, since Bayesian inference tends to  targeting the region with large probability mass, corresponding to the flat minima.

When analyzing SGD, most of existing works  assume the noise covariance of SGD is constant or upper bounded by some constant, and what role the noise structure of stochastic gradient plays in optimization and generalization was rarely studied in literature. 
On the other hand, experiments (Figure~\ref{fig:generalization_svhn_cifar}) show that the isotropic approximation of SGD like gradient Langevin dynamic (GLD) cannot fully explain the mystery of the good generalization performance of SGD, even they are tuned to have the same noise magnitude.
Thus the analysis over the structure of SGD noise is on demand for fully understanding SGD.
 
In this work, we take the first step studying the anisotropic noise of SGD and its superiority over its isotropic equivalence.
Specifically, we  study a general form of gradient-based optimization dynamics with unbiased noise, which unifies SGD and standard Langevin dynamics.
By investigating the general dynamics, we analyze how the noise structure of SGD influences the escaping behavior from sharp minima and its regularization effects.
Several novel  analysis and empirical justifications are made as follow.

(1) We derive a key indicator to characterize the efficiency of escaping from minima through measuring the alignment of noise covariance and the curvature of loss function.
Based on this indicator, two conditions are established to show which type of noise structure is superior to isotropic noise in term of escaping efficiency;

(2) We further justify that SGD in the context of neural networks satisfies these two conditions, and thus provide a plausible explanation why SGD can escape from sharp minima more efficiently, converging to flat minima with a higher probability. Moreover, these flat minima typically generalize well according to various works~\citep{hochreiter1997flat,keskar2016large,neyshabur2017,wu2017towards}. We also show that Langevin dynamics with well tuned isotropic noise cannot beat SGD, which further confirms the importance of noise structure of SGD;

(3) A large number of experiments are designed systematically to justify our understanding on the behavior of the anisotropic diffusion of SGD.
We compare SGD with full gradient descent with different types of diffusion noise, including isotropic and position-dependent/independent noise.
All these comparisons demonstrate the effectiveness of anisotropic diffusion for good generalization in training neural networks.


\begin{table*}
\centering
\caption{\small Compared dynamics defined in Eq.~(\ref{eq:gdnoise}).
The parameter $\sigma_t$ is adjusted to force the noise share the same expected norm as that of SGD noise, to meet constraint Eq.~\eqref{eq:tr-const} for fair comparison.}
\label{tb:dynamics}
\small
\begin{tabular}{|c|c|p{10cm}|}
\hline
Dynamics & Noise $\epsilon_t$ & Remarks\\ 
\hline
{\bf SGD} & $\epsilon_t \sim \Ncal \left(0, \Sigma^{\text{sgd}}_t\right)$ & $\Sigma^{\text{sgd}}_t$ is defined as in Eq.~(\ref{eq:sgdcov}). \\
\hline
{\bf GLD constant} & $\epsilon_t \sim \Ncal \left(0, \varrho_t^2 I\right)$ & $\varrho_t$ is a tunable constant. \\
\hline
{\bf GLD dynamic} & $\epsilon_t \sim \Ncal \left(0, \sigma_t^2 I\right)$ & $\sigma_t$ is adjusted to force the noise share the same magnitude with SGD noise, similarly hereinafter. \\
\hline
{\bf GLD diagonal} & $\epsilon_t \sim \Ncal \left(0, \diag(\Sigma_t^{\text{sgd}})\right)$ &  $\diag(\Sigma_t^{\text{sgd}})$ is the diagonal of the covariance of SGD noise $\Sigma_t^{\text{sgd}}$. \\
\hline
{\bf GLD leading} & $\epsilon_t \sim \Ncal \left(0, \sigma_t \tilde{\Sigma}_t\right)$ & $\tilde{\Sigma_t}$ is a low rank approximation of $\Sigma^{\text{sgd}}_t$, i.e., $\tilde{\Sigma}_t = \sum_{i=1}^k \gamma_i v_i v_i^T$, where  $\gamma_i, v_i$ are the first $k$ leading eigenvalues and corresponding unit eigenvectors of $\Sigma_t^{\text{sgd}}$. \\
\hline
{\bf GLD Hessian} & $\epsilon_t \sim \Ncal \left(0, \sigma_t \tilde{H}_t\right)$ & $\tilde{H}_t$ is a low rank approximation of the Hessian matrix of loss $L(\theta)$ by its the first $k$ leading eigenvalues and corresponding eigenvalues. \\
\hline
{\bf GLD 1st eigven($H$)} & $\epsilon_t \sim \Ncal \left(0, \sigma_t \lambda_1 u_1 u_1^T\right)$ & $\lambda_1, u_1$ are the maximal eigenvalue and its corresponding unit eigenvector of the Hessian matrix of loss $L(\theta_t)$. \\
\hline
\end{tabular}
\end{table*}

\section{Background}
\label{sec:back}

In general, supervised learning usually involves an optimization process of minimizing an empirical loss over training data,
\begin{equation}
    L(\theta) := \frac{1}{N} \sum_{i=1}^N \ell(x_i; \theta),
    \label{eq:model}
\end{equation}
where $\{x_i | i=1,\dots,N\}$ denotes the training set with $N$ \emph{i.i.d.} training samples, the model is parameterized by $\theta \in \Rbb^D $ and $\ell$ denotes the combination of the loss and the model for simplicity, e.g. deep networks with cross entropy loss.
Under many circumstances, including deep networks, there could exist multiple global minima for Eq.~(\ref{eq:model}), exhibiting diverse generalization performance.
We call those solutions generalizing well good solutions or minima, and vice versa.

\paragraph{Gradient descent and its stochastic variants}
A typical approach to minimize Eq.~(\ref{eq:model}) is gradient descent (GD),
\begin{equation}
    \theta_{t+1} = \theta_t - \eta \down_{\theta}L(\theta_t),
    \label{eq:gd}
\end{equation}
where $\eta$ denotes the learning rate and we assume it to be a small constant for the convenience of analysis, similarly hereinafter.

In practice, a more useful kind of gradient based optimizers act like GD with an unbiased noise, including gradient Langevin dynamics (GLD),
\begin{equation}
    \theta_{t+1} = \theta_t - \eta \down_{\theta}L(\theta_t) + \eta \epsilon_t, \epsilon_t \sim \Ncal \left(0, \sigma_t^2 I\right);
    \label{eq:gld}
\end{equation}
and stochastic gradient descent (SGD),
\begin{equation}
    \theta_{t+1} = \theta_t - \eta \tilde{g}(\theta_t),
    \label{eq:sgd0}
\end{equation}
where $\tilde{g}(\theta_t) = \frac{1}{m}\sum_{x \in B_t} \down_{\theta} \ell(x; \theta_t)$ is an unbiased estimator of the full gradient $\down_{\theta} L (\theta_t)$, with $B_t$ being a randomly selected minibatch of size $m$.
Assume the size of minibatch $m$ is large enough for the central limit theorem to hold, thus $\tilde{g}(\theta_t)$ follows a Gaussian distribution~\cite{chen2014stochastic,ahn2012bayesian,shang2015covariance,mandt2017stochastic},
\begin{equation}
    \begin{aligned}
        &\tilde{g} (\theta_t) \sim \Ncal \left(\down L(\theta_t), \Sigma^{\text{sgd}} (\theta_t) \right), \Sigma^{\text{sgd}}(\theta_t) \approx \\
        &\frac{1}{m} \left[\frac{1}{N}\sum_{i=1}^N\down\ell(x_i;\theta_t)\down\ell(x_i;\theta_t)^T - \down L(\theta_t) \down L(\theta_t)^T \right].
    \end{aligned}
    \label{eq:sgdcov}
\end{equation}
Therefore we can rewrite Eq.~(\ref{eq:sgd0}) as, 
\begin{equation}
    \theta_{t+1} = \theta_t - \eta \down L(\theta_t) + \eta \epsilon_t, \quad \epsilon_t \sim \Ncal \left(0, \Sigma^{\text{sgd}}(\theta_t) \right).
    \label{eq:sgd}
\end{equation}

Inspired by the dynamics of GLD (Eq.~(\ref{eq:gld})) and SGD (Eq.~(\ref{eq:sgd})), more generally, we study the dynamics of \emph{gradient descent with unbiased noise},
\begin{equation}
    \theta_{t+1} = \theta_t - \eta \down_{\theta}L(\theta_t) + \eta\epsilon_t, \quad \epsilon_t \sim \Ncal \left(0, \Sigma_t \right).
    \label{eq:gdnoise}
\end{equation}
For small enough constant learning rate $\eta$, Eq.~(\ref{eq:gdnoise})
can be treated as the numerical discretization of the following stochastic differential equation (SDE)
\citep{li2017stochastic,jastrzkebski2017three,chaudhari2017stochastic},
\begin{equation}
    \dif \theta_t = -\down_{\theta} L(\theta_t) \dif t  + \sqrt{\eta\Sigma_t}\dif W_t,
    \label{eq:sde}
\end{equation}
where $W_t$ is a standard Brownian motion in $\Rbb^D$.

Let $\Sigma_t=\Sigma^{\text{sgd}}(\theta_t)$ and $\sqrt{\eta\Sigma^{\text{sgd}}(\theta_t)}$ be the coefficient of the the noise term, \citet{hoffer2017} and \citet{jastrzkebski2017three} studied the generalization influence of the magnitude of the SGD noise, which is controlled by the quotient of learning rate and batch size, $\frac{\eta}{m}$.

Different from previous works either assuming the noise of SGD is constant or upper bounded by some constant, we are the first to study SGD from the perspective of its noise structure.
In the following sections,
we first show that for dynamics Eq.~(\ref{eq:sde}), the structure of $\Sigma_t$ indeed affects the escaping from minima, especially for the sharp ones containing rich curvature information; and then we demonstrate that for neural networks, the noise of SGD is closely related to the Hessian of loss surface.
Hence we conclude that SGD can escape from sharp minima much faster than its isotropic equivalence, and converge to flatter minima which tend to generalize better.
Finally we verify our understanding by numerous experiments.

\section{The behaviors of escaping from minima}
\label{sec:ou}

To ease the notation, we absorb $\eta$ into $\Sigma_t$ in Eq.~(\ref{eq:sde}), 
\begin{equation}
    \dif \theta_t = -\down_{\theta} L(\theta_t) \dif t  + \Sigma_t^{\half}\dif W_t.
    \label{eq:sde1}
\end{equation}

We now analyze the escaping behaviors of dynamics Eq.~\eqref{eq:sde1} with different choices of noise structures, i.e., $\Sigma_t$.

\subsection{The escaping efficiency}
We define the \emph{escaping efficiency} as the expected increase of the potential or the loss.
\begin{definition}[Escaping efficiency]
    Suppose we start the dynamics of Eq.~(\ref{eq:sde1}) from the minimum $\theta_0 $, then for a fixed time $t$ small enough (such that $L(\theta_t) - L(\theta_0) \ge 0$), we call 
    \begin{equation}
        \Ebb_{\theta_t}[L(\theta_t) - L(\theta_0)]
    \end{equation}
    the \emph{escaping efficiency}.
\end{definition}
There are two remarks about the definition of escaping efficiency.
Firstly it characterizes the ability of the dynamic escaping from the minimum $\theta_0$.
Secondly because $L(\theta_t) - L(\theta_0) \ge 0$, for any $\delta > 0$, the escaping probability $P(L(\theta_t)-L(\theta_0) \ge \delta)$ can be upper bounded by the expectation $\Ebb[L(\theta_t) - L(\theta_0)]$, given the Markov's inequality, $P(L(\theta_t) - L(\theta_0) \ge \delta) \le \frac{\Ebb [L(\theta_t)-L(\theta_0)]}{\delta}$.

Now we calculate the escape efficiency of dynamics Eq.~(\ref{eq:sde1}).
Provided that the mild smoothness assumptions for Ito's lemma holds, we have
\begin{equation}
    \Ebb [L(\theta_t)-L(\theta_0)] = - \int_{0}^{t} \Ebb \left[ \down L^T \down L\right]+\int_{0}^{t} \half \Ebb \Tr (H_t\Sigma_t) \dif t,
    \label{eq:escape_sde}
\end{equation}
where $H_t := \down^2_\theta L(\theta_t)$ is the Hessian of $L(\theta_t)$.
The derivation of Eq.~\eqref{eq:escape_sde} is provided in Supplementary Materials.

Generally, the escaping efficiency characterized by Eq.~\eqref{eq:escape_sde} is hard to analyze due to the intractableness of the integral.
Nonetheless, focusing on the locally escaping process, we take the second-order approximation near the minima $\theta_0$, where $L(\theta) \approx L(\theta_0) + \half (\theta - \theta_0)^T H (\theta - \theta_0)$.
Without loss of generality, let $\theta_0 = 0$.
Further, assume $H$ is a positive definite matrix and the diffusion covariance matrix $\Sigma_t=\Sigma$ is constant for $t$.
Then Eq.~\eqref{eq:sde1} becomes an Ornstein-Uhlenbeck process,
\begin{equation}
    \dif \theta_t = -H \theta_t \dif t + \Sigma^{\half} \dif W_t, \quad \theta_0 = 0. \label{eq:ou}
\end{equation}

The escaping efficiency of Eq.~\eqref{eq:ou} could be explicitly obtained as
\begin{equation}
    \Ebb [L(\theta_t)-L(\theta_0)] = \frac{1}{4} \Tr\left(\left( I-e^{-2Ht}\right)\Sigma\right)\approx \frac{t}{2} \Tr\left(H\Sigma\right).
    \label{eq:escape_ou}
\end{equation}
We defer the derivation to Supplementary Materials.

Eq.~(\ref{eq:escape_sde}) and Eq.~(\ref{eq:escape_ou}) characterize the escaping efficiency of general process and  Ornstein-Uhlenbeck process respectively, and they clearly show that the indicator $\Tr(H_t\Sigma_t)$ plays an crucial role for stochastic processes escaping from minima.
Since we only care about the locally escaping behavior near the minima, we could directly analyze this key indicator $\Tr (H_t\Sigma_t)$, in order to understand the importance of noise structure $\Sigma_t$ for escaping.

\subsection{Anisotropic noise helps escape from sharp minima}
Now we study what factors affect the locally escaping behaviors by analyzing the indicator $\Tr (H_t\Sigma_t)$.

\paragraph{The magnitude of noise}
Clearly, the magnitude of the noise affects the escaping efficiency and larger magnitude leads to faster escape.
Along this line, \citet{hoffer2017} and \citet{jastrzkebski2017three} studied the generalization influence of the magnitude of the SGD noise, which is controlled by the quotient of learning rate and batch size.

Hence to explore the role of the noise structure, we must eliminate the impact of noise magnitude for fair comparison.
One reasonable evaluation of the noise magnitude is the expected squared norm of the noise vector~\citep{li2017stochastic}:
suppose $\epsilon_t \sim \Ncal(0, \Sigma_t), z\sim\Ncal(0,I)$ and the eigen decomposition of $\Sigma_t$ is $\Sigma_t = V \Gamma V^T$, then
\begin{equation*}
    \begin{aligned}
        \norm{\epsilon_t}_{\text{trace}} &:= \Ebb [\epsilon_t^T \epsilon_t] = \Ebb [(V\sqrt{\Gamma}z)^T (V\sqrt{\Gamma}z)] = \Ebb [z^T \Gamma z] \\
        &= \Ebb \Tr (\Gamma z z^T) = \Tr \Ebb [\Gamma z z^T] = \Tr (\Sigma_t).
    \end{aligned}
\end{equation*}
Based on such measure of magnitude, we introduce the following important trace constraint,
\begin{equation}
    \textbf{given time } t,  \Tr (\Sigma_t)  \textbf{ is constant}.
    \label{eq:tr-const}
\end{equation}
From the statistical physics point of view, $\Tr (\Sigma_t)$ characterizes the kinetic energy~\cite{gardiner2009stochastic}, thus it is natural to force the energy to be unchanging, otherwise it is trivial that the higher the energy is, the less stable the system is.

\paragraph{The ill-conditioning of minima}
Consider the isotropic minima where the Hessian is $H_t=\lambda I$, our escaping indicator becomes $\Tr(H_t\Sigma_t) = \lambda \Tr\Sigma_t$, which is invariant under constraint Eq.~\eqref{eq:tr-const}.
Thus the noise structure has no impact on escaping from isotropic minima.
However, for the minima where the Hessian is highly ill-conditioned, which is the typical case in practical over-parameterized neural networks~\cite{sagun2017empirical}, the noise structure could cause huge difference on escaping behaviors, as analyzed below. 

\paragraph{The structure of noise}
For semi-positive definite $H_t, \Sigma_t$ and assuming $H_t$ has distinguished top eigenvalues, to achieve the maximum of $\Tr(H_t\Sigma_t)$ under constraint Eq.\eqref{eq:tr-const}, $\Sigma_t$ should be $\Sigma_t^* = (\Tr\Sigma_t)\cdot u_1 u_1^T$, where $u_1$ is the first unit eigenvector of $H_t$.
Note this rank-1 matrix $\Sigma_t^*$ is highly anisotropic.
More generally, the following Proposition~\ref{thm:aniso_benefit} characterizes one kind of anisotropic noise significantly outperforming its isotropic equivalence, given $H$ is ill-conditioned.

\begin{proposition}[The benefits of anisotropic noise]
Assume $H_{D\times D}$ and $\Sigma_{D\times D}$ are semi-positive definite. If 

(1) $H$ is ill-conditioned.
Let $\lambda_1 \ge \lambda_2 \ge \dots, \ge \lambda_D \ge 0$ be the eigenvalues of $H$ in descent order, and for some constant $k\ll D$ and $d>\half$,
\begin{equation}
    \lambda_1 > 0,\qquad \lambda_{k+1}, \lambda_{k+2}, \dots, \lambda_{D} < \lambda_1 D^{-d};
    \label{eq:cond1_H_ill}
\end{equation}
(2) $\Sigma$ is ``aligned'' with $H$.
Let $u_i$ be the corresponding unit eigenvector of eigenvalue $\lambda_i$, for some projection coefficient $a>0$,
\begin{equation}
    u_1^T\Sigma u_1 \ge a \lambda_1 \frac{\Tr\Sigma}{\Tr H};
    \label{eq:cond2_Sigma_align}
\end{equation}
then for such $\Sigma$ and its isotropic equivalence $\bar{\Sigma} = \frac{\Tr\Sigma}{D}I$ under constraint Eq.~(\ref{eq:tr-const}),
we have the follow ratio describing their difference in term of escaping efficiency,
\begin{equation}
    \frac{\Tr\left(H\Sigma\right)}{\Tr(H\bar{\Sigma})} = \Ocal \left(a D^{(2d-1)}\right), \quad d > \frac{1}{2}.
    \label{eq:ratio}
\end{equation}
\label{thm:aniso_benefit}
\end{proposition}
The first condition Eq.~\eqref{eq:cond1_H_ill} characterizes the illness of $H$.
To give some geometric intuitions on the second condition Eq.~(\ref{eq:cond2_Sigma_align}), let the maximal eigenvalue and its corresponding unit eigenvector of $\Sigma$ be $\gamma_1, v_1$, then
$ u_1^T\Sigma u_1 \geq u_1^Tv_1\gamma_1 v_1^T u_1 = \gamma_1 \left<u_1, v_1\right>^2 $.
Thus if the maximal eigenvalues of $H$ and $\Sigma$ are aligned in proportion, $\gamma_1/\Tr\Sigma \ge a_1 \lambda_1/\Tr H$, and the angle of their corresponding unit eigenvectors is close enough such that $\left<u_1, v_1\right> \ge a_2$, the second condition Eq.~(\ref{eq:cond2_Sigma_align}) holds for $a =a_1 a_2$.

Typically, in the scenario of modern neural networks, due to the over-parameterization, Hessian and the gradient covariance are usually ill-conditioned and anisotropic near minima~\citep{sagun2017empirical,chaudhari2017stochastic}. Thus the first condition in Proposition~\ref{thm:aniso_benefit} usually holds for neural networks, and we further justify it by experiments in Section~\ref{sec:verification}. 
In the next section, we turn to discuss how the covariance of SGD noise meets the second condition of Proposition~\ref{thm:aniso_benefit} in the context of neural networks.
Hence this explains the superiority of the anisotropic noise of SGD over the isotropic one, such as gradient Langevin dynamics. 

\section{The relationship between the noise of SGD and the curvature of loss surface}
\label{sec:sgdnoise}

In this section we investigate the anisotropic structure of gradient covariance in SGD, and explore its connection with the Hessian of loss surface.

\textbf{Around the true parameter.} 
According to the classic statistical theory~\citep[Chap. 8]{pawitan2001all}, for population loss $L(\theta) = \Ebb_x \ell(x; \theta)$, with $\ell$ being the negative log likelihood, when evaluating at the true parameter $\theta^*$,
there is the exact equivalence between the Hessian $H$ of the population loss and \emph{ Fisher information matrix} $F$ at $\theta^*$,
\begin{equation*}
\begin{aligned}
    F(\theta^*) &:= \Ebb_x [\down_{\theta} \ell(x; \theta^*) \down_{\theta} \ell(x; \theta^*)^T] = \Ebb_x [\nabla_{\theta}^2 \ell(x; \theta^*)]\\
    &= \down^2_{\theta} L(\theta^*) =: H(\theta^*).
\end{aligned}
\end{equation*}


In practice, with the assumptions that the sample size $N$ is large enough (i.e. indicating asymptotic behavior) and suitable smoothness conditions, when the current parameter $\theta_t$ is not far from the ground truth, Fisher is close to Hessian. Thus we can obtain the following approximate equality between gradient covariance and Hessian,
\begin{equation}
    \begin{aligned}
            \hat{\Sigma} (\theta_t) &= \hat{F} (\theta_t) - \down_{\theta_t} \hat{L}(\theta_t)\down_{\theta}\hat{L}^T(\theta_t) \approx \hat{F} (\theta_t) \approx \hat{F} (\theta^*) \\ &\approx F(\theta^*) = H(\theta^*)  \approx \hat{H} (\theta^*) \approx \hat{H}(\theta_t).
    \end{aligned}
\end{equation}
The first approximation is due to the dominance of noise over the mean of gradient in the later stage of SGD optimization~\citep{shwartz2017opening}, in which 
a similar experiment was conducted to demonstrate this observation, shown in Supplementary Materials due to the limit of space.

\textbf{One hidden layer network with fixed output layer.} 
In the following we provide theoretical characterization about the alignment between $\Sigma$ and $H$ in the context of one hidden layer neural networks with fixed output layer.
We first show the connection of Fisher and Hessian in this specific case.

\begin{proposition}[The connection between Fisher and Hessian in one hidden layer network]
Consider a binary classification problem with data $\{(x_i, y_i)\}_{i\in I}, y\in\{0,1\}$, and mean square loss (either population or empirical),
\begin{equation}
    L(\theta) = \Ebb_{(x,y)}\norm{ \phi \circ f(x; \theta) - y}^2.
\end{equation}
Here $f$ denotes the network and $\phi$ is a threshold activation function controlling the output of the model,
\begin{equation}
    \phi (f) = \min \{\max \{f, \delta \}, 1-\delta\} \subset [\delta, 1-\delta],
\end{equation}
where $\delta$ is a small positive constant.

Suppose the network $f$ satisfies:
(1) it has one hidden layer and piece-wise linear activation;
(2) the parameters of its output layer are fixed during training~\cite{brutzkus2017sgd}.

Then for Fisher $F$ and Hessian $H$ (either population or empirical), we have

(1) $F(\theta) \succeq \delta ^2 H(\theta)$, almost everywhere; 
(2) $F(\theta) \preceq (\delta+\epsilon)^2 H(\theta)$, almost everywhere around the minima, $\{\theta : \norm{\phi \circ f(x;\theta)-y} \le \delta + \epsilon, \forall (x,y) \}$. 
$A\preceq B$ means that $(B-A)$ is semi-positive definite.
\label{thm:F_H}
\end{proposition}

There are two remarks on Proposition~\ref{thm:F_H}.
Firstly, the considered neural networks in Proposition~\ref{thm:F_H} are non-convex and have multiple minima, and one example to show this is provided in Supplementary Materials. Thus it is non-trivial to consider the escaping from minima.
Secondly, Proposition~\ref{thm:F_H} holds in both population and empirical sense, since the proof does not distinguish the two circumstances.

Based on Proposition~\ref{thm:F_H}, we could show that this neural network meets the second condition in Proposition~\ref{thm:aniso_benefit}.

\begin{proposition}[The connection between gradient covariance and Hessian in one hidden layer network]
Assume the conditions in Proposition~\ref{thm:F_H} hold, then there is a constant $a >0$, for $\theta$ close enough to minima $\theta^*$ (local or global), we have
\begin{equation}
    u(\theta)^T \Sigma(\theta) u(\theta) \ge a \lambda(\theta) \frac{\Tr \Sigma(\theta)}{\Tr H(\theta)}
\end{equation}
holds almost everywhere, for $\lambda(\theta)$ and $u(\theta)$ being the maximal eigenvalue and its corresponding eigenvector of Hessian $H(\theta)$.
\label{thm:Sigma_H}
\end{proposition}

Therefore, based on the discussion on population loss around the true parameters and one hidden layer neural network with fixed output layer parameters, given the ill-conditioning of $H$ due to the over-parameterization of modern neural networks, according to Proposition~\ref{thm:aniso_benefit}, we can conclude the noise structure of SGD helps to escape from sharp minima significantly faster than the dynamics with isotropic noise.
Hence SGD tends to converge to flatter solutions, which typically generalize well~\citep{hochreiter1997flat,keskar2016large,neyshabur2017,wu2017towards}.
Thus, the anisotropic noise of SGD might explain its better generalization performance comparing to GD, GLD and other dynamics with isotropic noise.

In the following, we conduct a series of experiments systematically to verify our understanding on the behavior of escaping from minima and its regularization effects for different optimization dynamics.

\begin{figure*}
\centering
\begin{tabular}{ccc}
\includegraphics[width=0.3\linewidth]{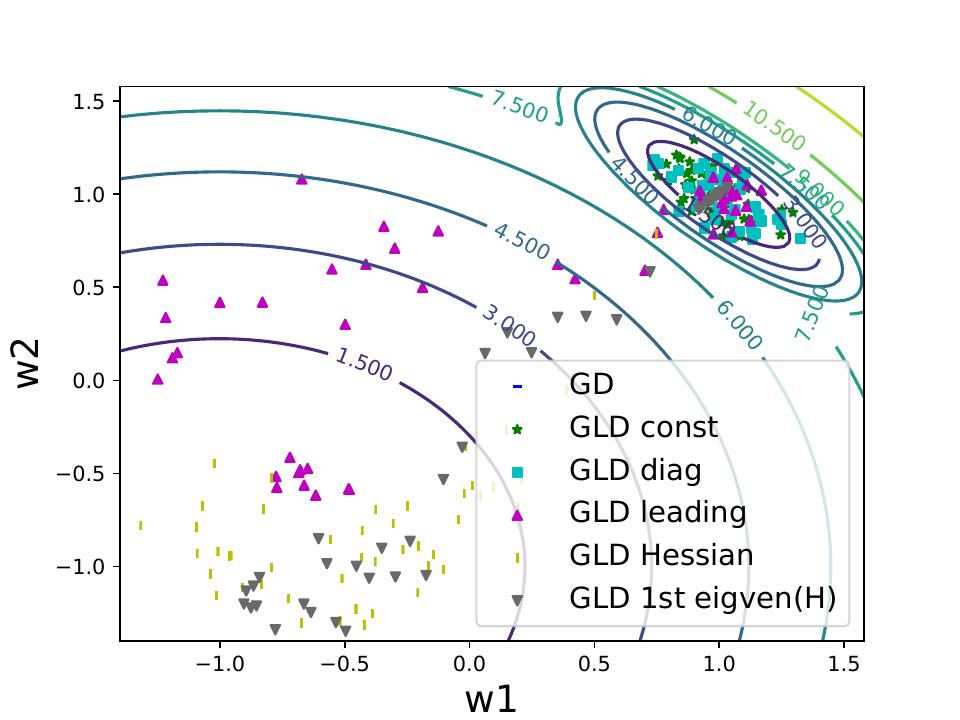} & \includegraphics[width=0.3\linewidth]{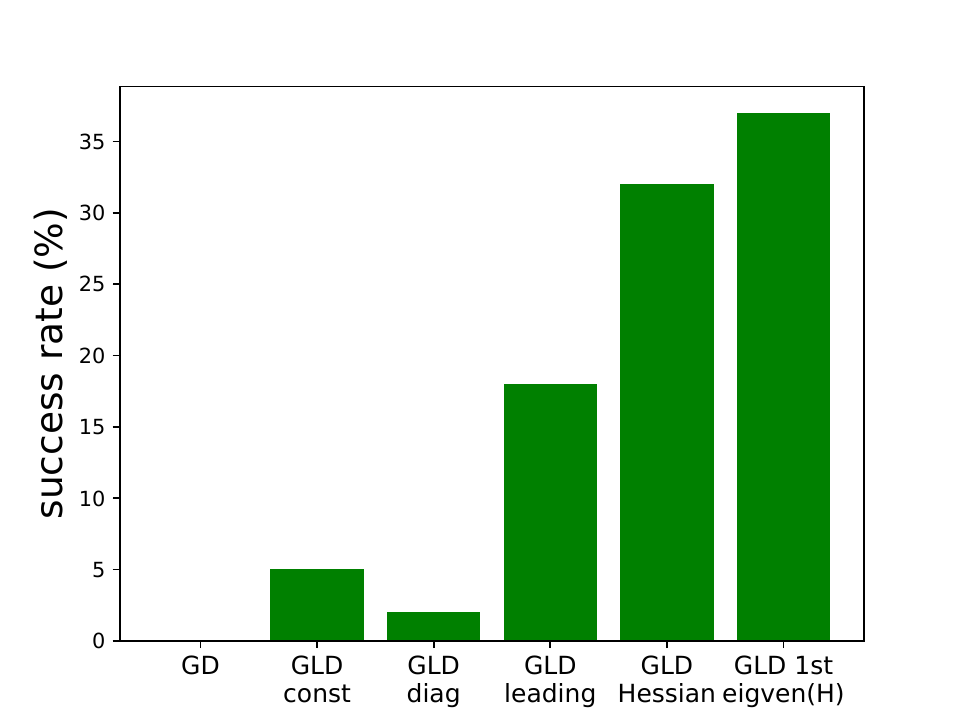} & \includegraphics[width=0.3\linewidth]{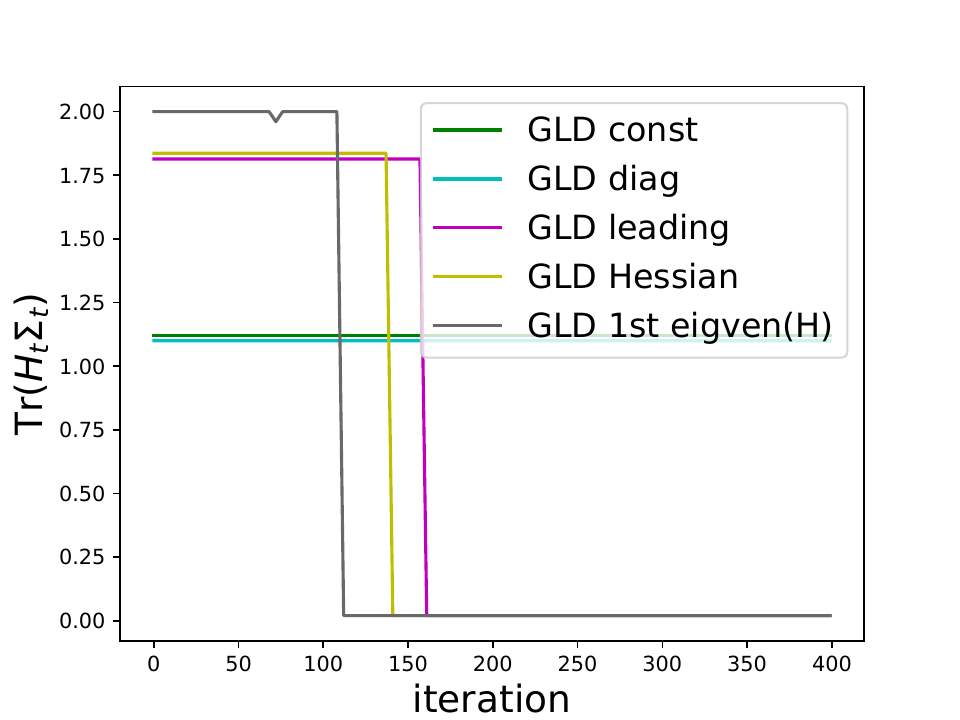}
\end{tabular}
\vspace{-4mm}
\caption{\small 2-D toy example. Compared dynamics are defined in Table~\ref{tb:dynamics}, $k=2$, $\sigma_t^2$ is tuned to keep noise of all dynamics sharing same expected squared norm, $0.01$.
All dynamics are run by $500$ iterations with learning rate $0.005$.
\textbf{Left}: The trajectory of each compared dynamics for escaping from the sharp minimum in one run.
\textbf{Middle}: Success rate of arriving the flat solution in $100$ repeated runs.
\textbf{Right}: $\Tr(H_t \Sigma_t)$ of compared dynamics in one run.}
\label{fig:2d}
\vspace{-4mm}
\end{figure*}

\begin{figure*}
\centering
\begin{tabular}{ccc}
    \includegraphics[width=0.3\linewidth]{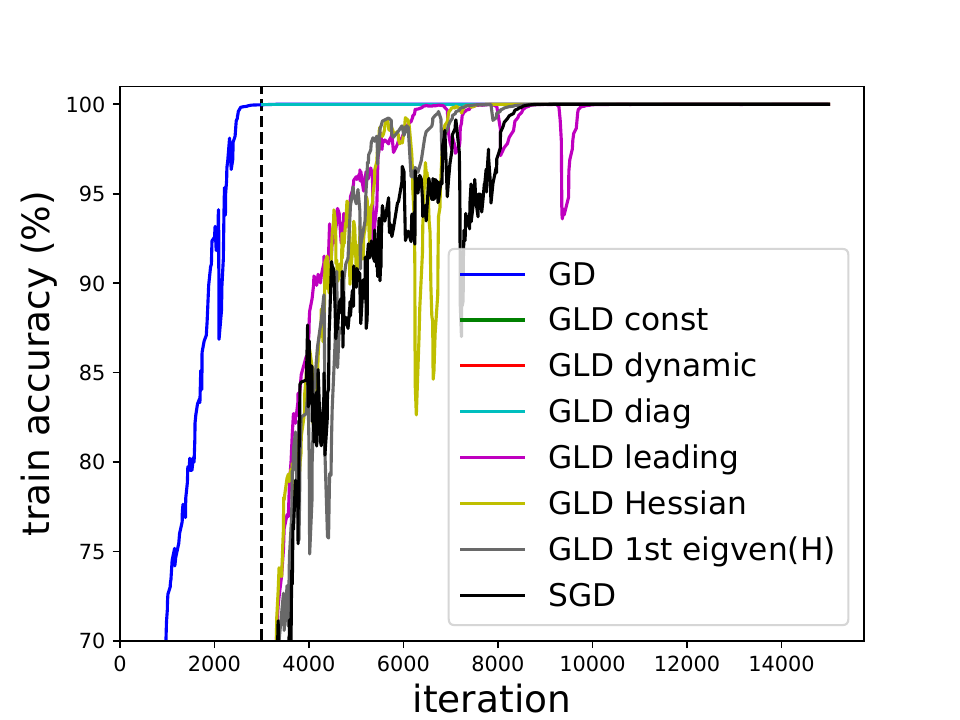} &  \includegraphics[width=0.3\linewidth]{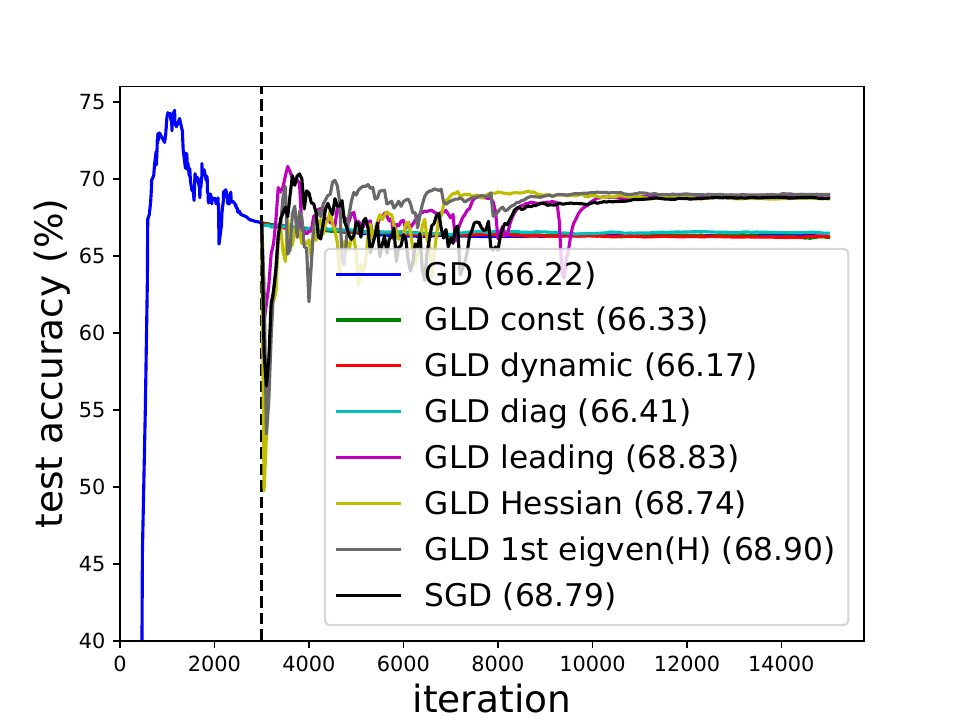} & \includegraphics[width=0.3\linewidth]{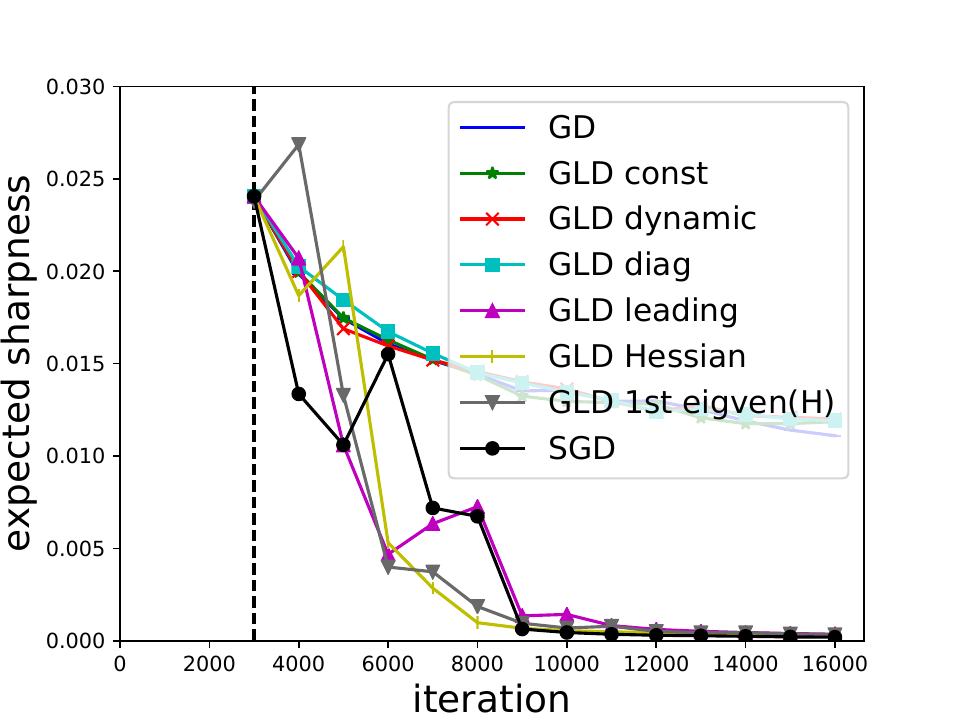}
\end{tabular}
\vspace{-4mm}
\caption{\small FashionMNIST experiments. Compared dynamics are initialized at $\theta^*_{GD}$ found by GD, marked by the vertical dashed line in iteration $3000$. The learning rate is same for all the compared methods, $\eta_t = 0.07$, and batch size $m=20$.
\textbf{Left}: Training accuracy versus iteration.
\textbf{Middle}: Test accuracy versus iteration. The final accuracy is noted within the parentheses.
\textbf{Right}: Expected sharpness versus iteration. Expected sharpness is measured as $\Ebb_{\nu \sim \Ncal(0, \delta^2 I)} \left[ L(\theta + \nu)\right] - L(\theta)$, and $\delta=0.01$, the expectation is computed by average on $1000$ times sampling.}
\label{fig:fashion}
\vspace{-4mm}
\end{figure*}

\section{Experiments}
\label{sec:exp}

For better understanding the difference between the anisotropic noise and the isotropic one, we introduce dynamics with various kinds of noise structure to empirical study with, as shown in Table~\ref{tb:dynamics}.

\subsection{Two-dimensional toy example}
We design a 2-D toy example  $L(w_1, w_2)$ with two basins, a small one and a large one, corresponding to a sharp and flat minima, $(1,1)$ and $(-1,-1)$, respectively, both of which are global minima, see Supplementary Materials for more details.
We initialize the dynamics of interest with the sharp minimum $(w_1, w_2) = (1,1)$, and run them to study their behaviors escaping from this sharp minimum.

To explicitly control the noise magnitude, we only conduct experiments on GD, GLD const, GLD diag, GLD leading (with $k=2=D$ in Table~\ref{tb:dynamics}, which is also the exactly covariance of SGD noise), GLD Hessian ($k=2$) and GLD 1st eigen($H$).
And we adjust $\sigma_t$ in each dynamics to force their noise to share the same expected squared norm the meet the constraint Eq.~\eqref{eq:tr-const}.
Figure~\ref{fig:2d} (Left) shows the trajectories of the dynamics escaping from the sharp minimum $(1,1)$ towards the flat one $(-1,-1)$, while Figure~\ref{fig:2d} (Middle) presents the success rate of escaping for each dynamic during $100$ repeated experiments. Figure~\ref{fig:2d} (Right) demonstrates our derived indicator $\Tr(H_t\Sigma_t)$ in one run.

As shown in Figure~\ref{fig:2d}, GLD 1st eigvec($H$) achieves the highest success rate, indicating the fastest escaping speed from the sharp minimum.
The dynamics with anisotropic noise aligned with Hessian well, including GLD 1st eigvec($H$), GLD Hessian and GLD leading, greatly outperform GD, GLD const with isotropic noise, and GLD diag with noise poorly aligned with Hessian.
These experiments are consistent with our theoretical analysis on OU process shown in Eq.~\eqref{eq:escape_ou} and Proposition~\ref{thm:aniso_benefit}, demonstrating the benefits of anisotropic noise for escaping from sharp minima.

\subsection{One hidden layer network with fixed output layer}
\begin{figure}
\centering
\includegraphics[width=0.6\columnwidth]{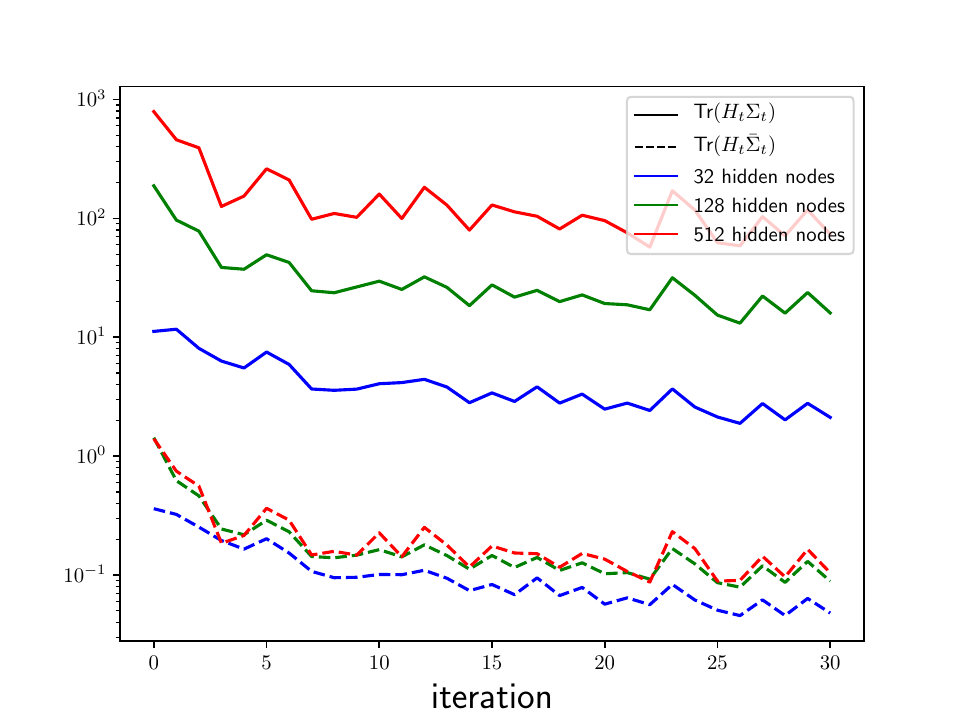}
\vspace{-4mm}
\caption{\small One hidden layer neural networks.
The solid and the dotted lines represent the value of $\Tr(H\Sigma)$ and $\Tr(H\bar{\Sigma})$, respectively.
The number of hidden nodes varies in $\{32, 128, 512\}$.} 
\label{fig:onehiddenlayer}
\vspace{-4mm}
\end{figure}

To verify the conclusion of Proposition~\ref{thm:aniso_benefit} in neural network cases, 
three networks are trained to binary classify $1,000$ linearly separable two-dimensional points to show the benefits of anisotropic noise of SGD. 
The activations are all ReLU and $\delta$ (in Proposition~\ref{thm:F_H}) is set to be $0.001$.
The number of hidden nodes for each network varies in $\{32,128,512\}$.
We plot the empirical indicator $\Tr \left(H\Sigma\right)$ in Figure~\ref{fig:onehiddenlayer}.
We can easily observe that as the increase of the number of hidden nodes, the ratio ${\Tr \left(H\Sigma\right)}/{\Tr \left(H\bar{\Sigma}\right)}$ is enlarged significantly, which is consistent with the Eq.~(\ref{eq:ratio}) described in Proposition~\ref{thm:aniso_benefit}.

\vskip -3mm
\subsection{FashionMNIST with corrupted labels}
We conduct a series of experiments in real deep learning scenarios to study the importance of SGD's noise covariance structure and its implicit regularization effects. 
We construct a noisy training set based on FashionMNIST dataset.
Concretely, the training set consist of $1000$ images with correct labels, and another $200$ images with random labels. 
A small LeNet-like network with $11,330$ parameters is utilized such that the spectral decomposition over  $\Sigma$ and $H$ are computationally feasible.

We firstly run the full gradient decent for $3,000$ iterations to arrive at the parameters $\theta^*_{GD}$ near the global minima with nearly zero training loss and $100\%$ training accuracy, which are typically sharp minima that generalize poorly~\citep{neyshabur2017}.
And then all other compared methods are initialized with $\theta^*_{GD}$ and run with the same learning rate $\eta_t = 0.07$ and same batch size $m=20$ (if needed) for fair comparison.

\textbf{Behaviors of different dynamics escaping from minima and its generalization effects.}
To compare the different dynamics on escaping behaviors and  generalization performance, we run dynamics initialized from the sharp minima $\theta^*_{GD}$ found by GD.
The settings for each compared method are as follows. The hyperparameter $\sigma^2$ for GLD const has already been tuned as optimal ($\sigma=0.001$) by grid search. For GLD leading, we set $k=20$ for comprising the computational cost and approximation accuracy. As for GLD Hessian, to reduce the expensive evaluation of such a huge Hessian in each iteration, we set $k=20$ and update the Hessian every $10$ iterations. We adjust $\sigma_t$ in GLD dynamic, GLD Hessian and GLD 1st eigvec($H$) to guarantee that they share the same expected squred noise norm defined in Eq.~(\ref{eq:tr-const}) as that of SGD.
And we measure the expected sharpness of different minima as $\Ebb_{\nu \sim \Ncal(0, \delta^2 I)} \left[ L(\theta + \nu)\right] - L(\theta)$, as defined in (\citep{neyshabur2017}, Eq.(7)). 

As shown in Figure~\ref{fig:fashion}, SGD, GLD 1st eigvec($H$), GLD leading and GLD Hessian successfully escape from the sharp minima found by GD, while GLD, GLD dynamic and GLD diag are trapped in the minima.
This demonstrates that the methods with anisotropic noise ``aligned'' with loss curvature can help to find flatter minima that generalize well. 

\textbf{Verification of the conditions in Proposition~\ref{thm:aniso_benefit}.} \label{sec:verification}
\begin{figure}
\centering
\begin{tabular}{ccc}
    \includegraphics[width=0.31\linewidth]{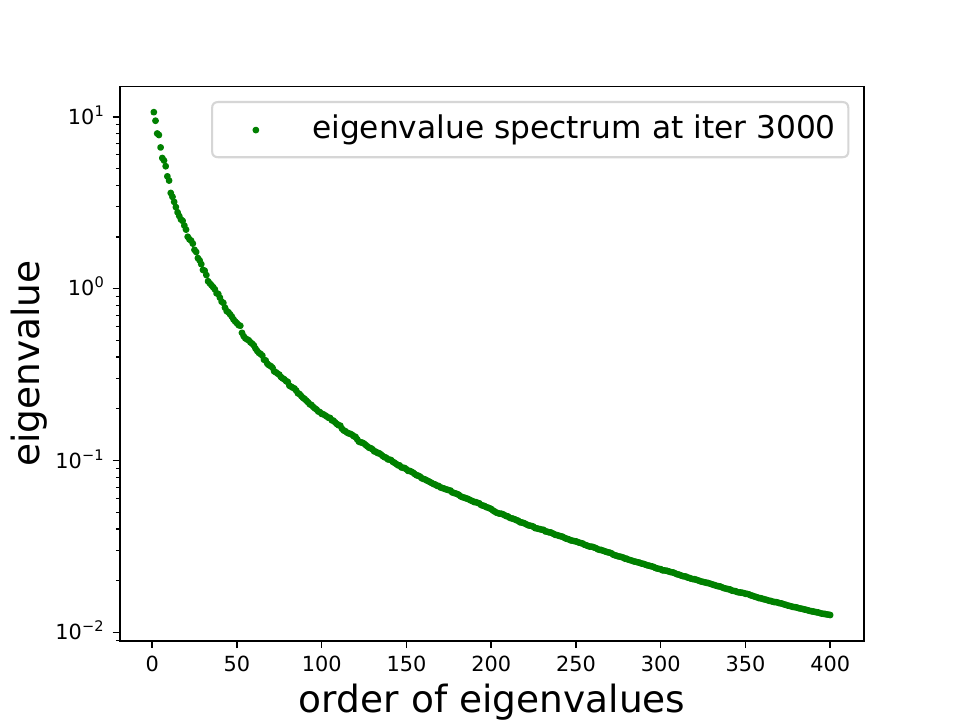} & \includegraphics[width=0.31\linewidth]{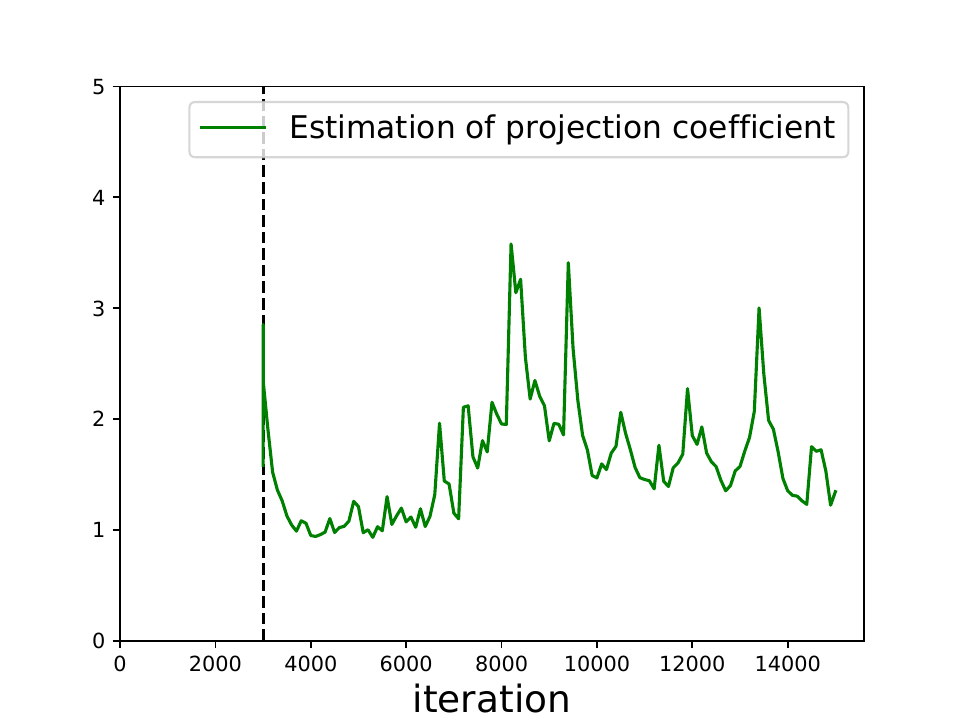} &  \includegraphics[width=0.31\linewidth]{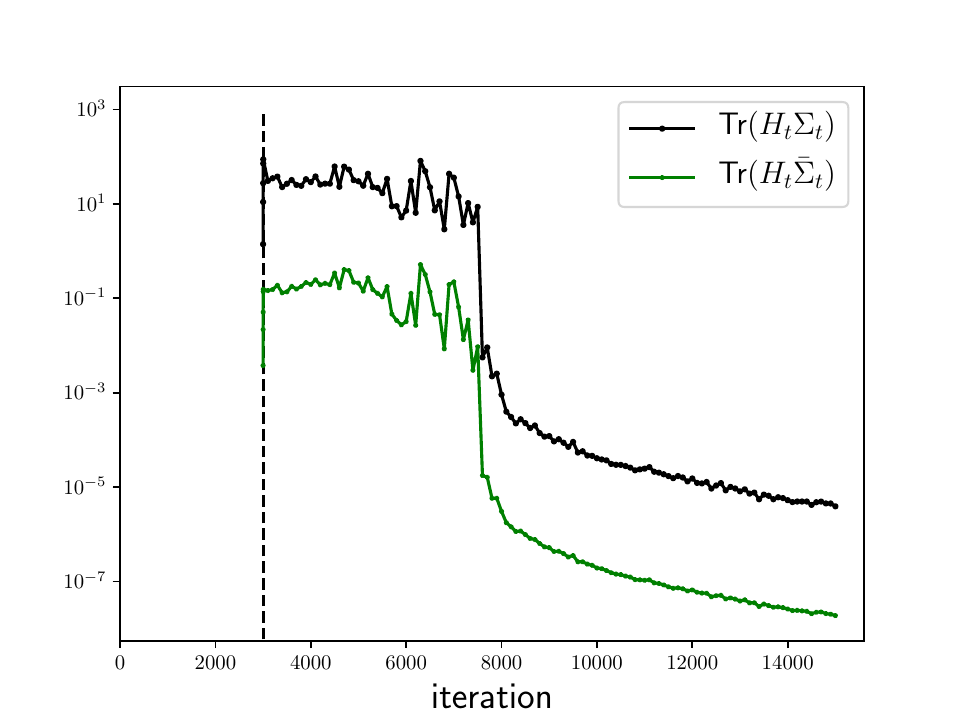}
\end{tabular}
\vspace{-4mm}
\caption{\small FashionMNIST experiments.
\textbf{Left}: The first $400$ eigenvalues of Hessian at $\theta^*_{GD}$, the sharp minima found by GD after $3000$ iterations.
\textbf{Middle}: The projection coefficient estimation $\hat{a} = \frac{u_1^T \Sigma u_1 \Tr H}{\lambda_1 \Tr\Sigma}$, as shown in Proposition~\ref{thm:aniso_benefit}.
\textbf{Right}: $\Tr(H_t\Sigma_t)$ versus $\Tr(H_t\bar{\Sigma}_t)$ during SGD optimization initialized from $\theta^*_{GD}$, $\bar{\Sigma}_t = \frac{\Tr\Sigma_t}{D}I$ denotes the isotropic noise with same expected squared norm as SGD noise.}
\label{fig:sgdnoise}
\end{figure}

To check whether the noise of SGD in deep neural networks  satisfies the two conditions in Proposition~\ref{thm:aniso_benefit}, we run SGD  initialized from $\theta^*_{GD}$, i.e. the sharp minima found by GD.

\begin{figure*}
\centering
\begin{tabular}{ccc}
    \includegraphics[width=0.3\linewidth]{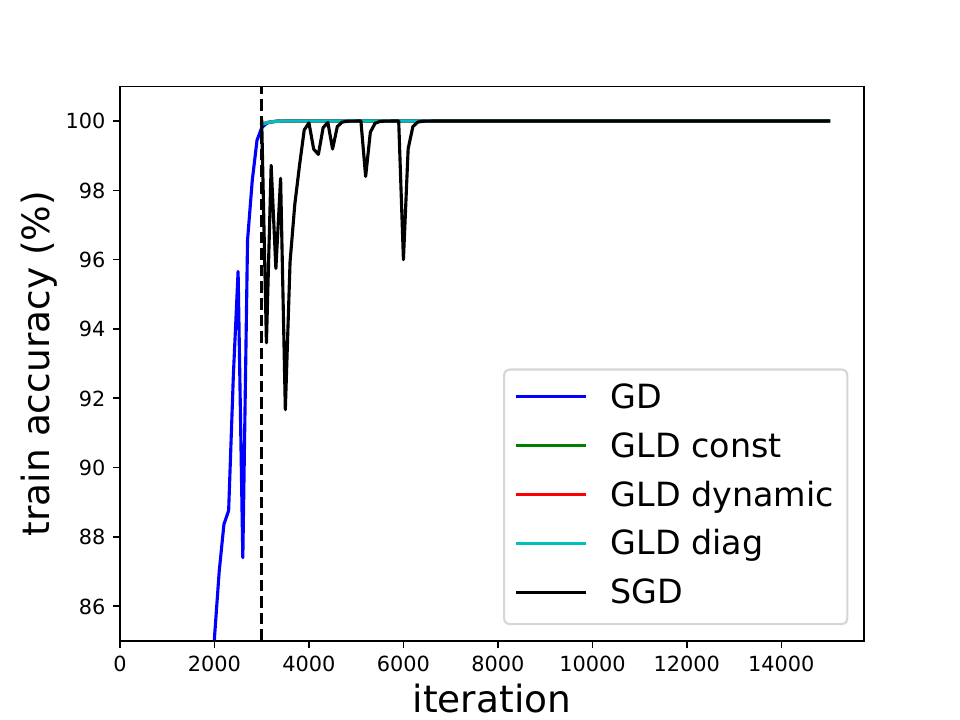} &  \includegraphics[width=0.3\linewidth]{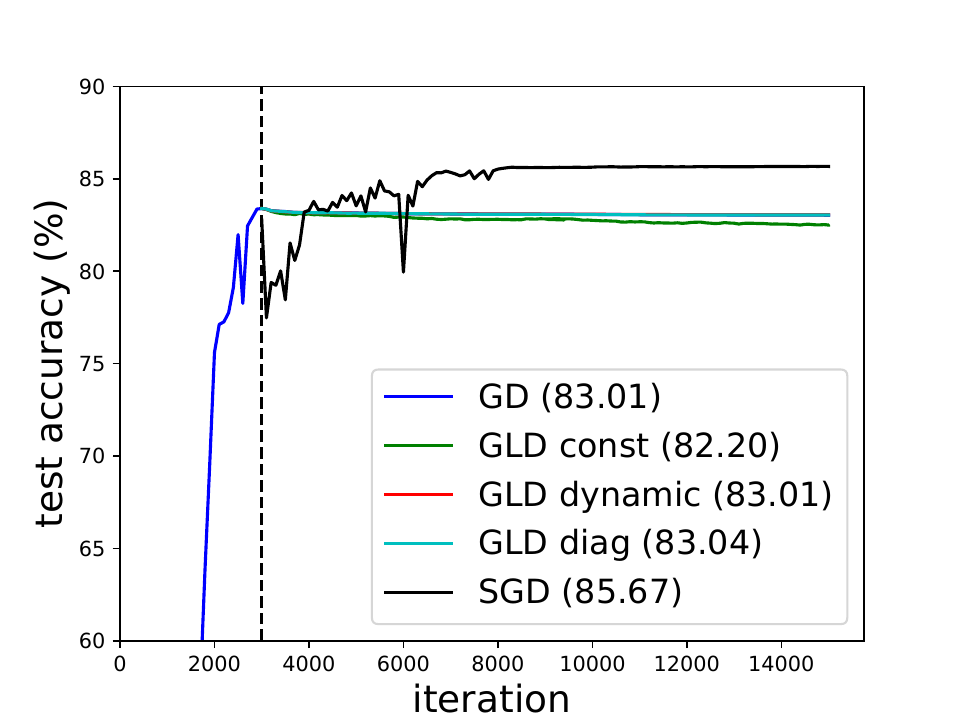} & \includegraphics[width=0.3\linewidth]{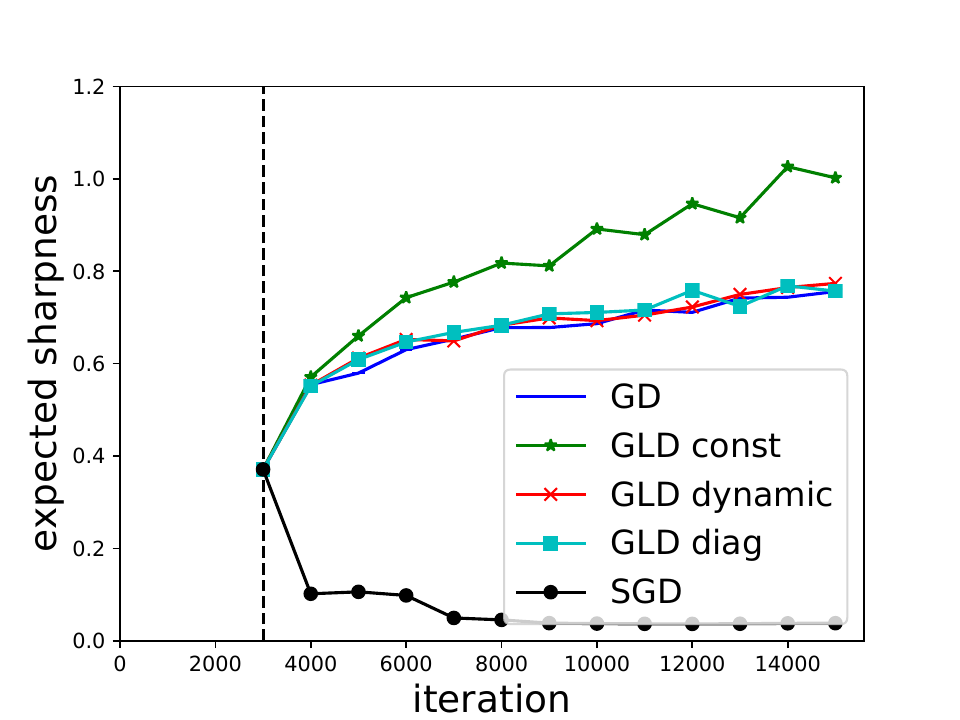}\\
    \includegraphics[width=0.3\linewidth]{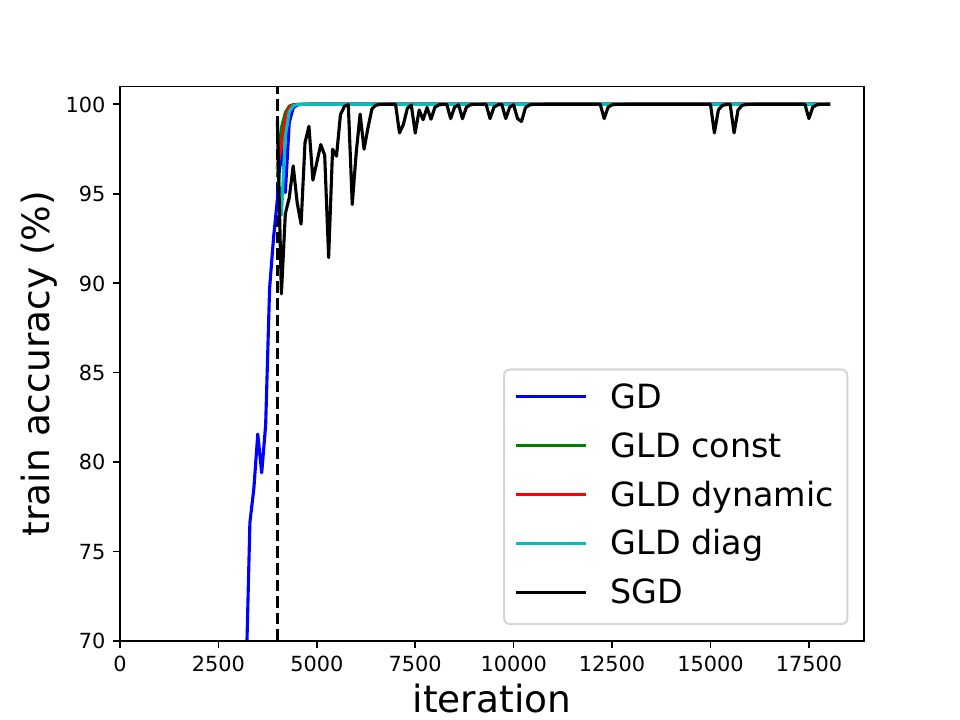} &  \includegraphics[width=0.3\linewidth]{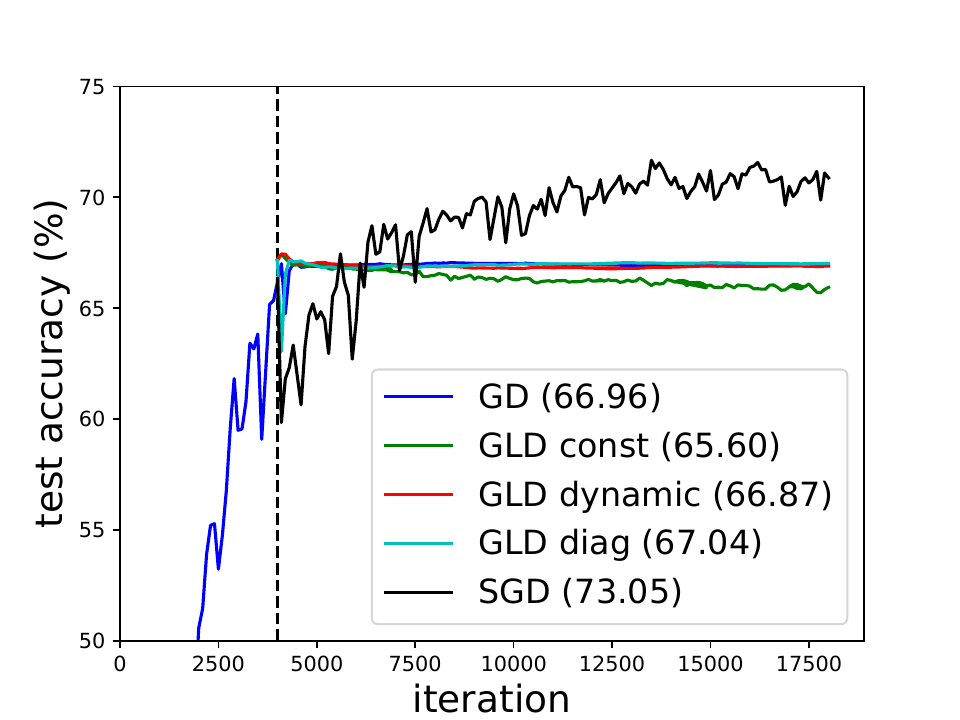} & \includegraphics[width=0.3\linewidth]{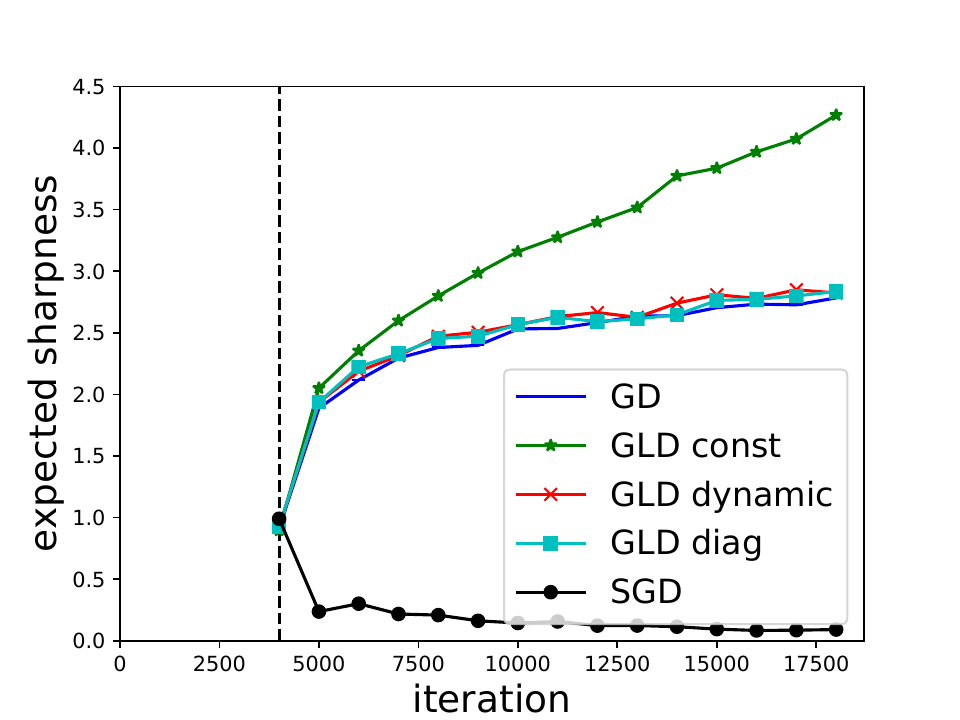}\\
\end{tabular}
\vspace{-4mm}
\caption{\small SVHN and CIFAR-10 experiments. 
\textbf{Top}: SVHN experiments;
\textbf{Bottom}: CIFAR-10 experiments.
Compared dynamics are initialized at $\theta^*_{GD}$ found by GD, marked by the vertical dashed line (in iteration $3,000$ for SVHN and iteration $4000$ for CIFAR-10).
The learning rate is same for all the compared methods, $\eta_t = 0.05$, and batch size $m=100$.
\textbf{Left}: Training accuracy versus iteration.
\textbf{Middle}: Test accuracy versus iteration. The final accuracy is noted within the parentheses.
\textbf{Right}: Expected sharpness versus iteration.
Expected sharpness is measured as $\Ebb_{\nu \sim \Ncal(0, \delta^2 I)} \left[ L(\theta + \nu)\right] - L(\theta)$, and $\delta=0.01$, the expectation is computed by average on $100$ times sampling.}
\label{fig:escape_svhn_cifar}
\vspace{-4mm}
\end{figure*}

Figure~\ref{fig:sgdnoise}(Left) shows the first $400$ eigenvalues of Hessian at $\theta^*_{GD}$, from which we see that the $140$th eigenvalue has already decayed to about $1\%$ of the first eigenvalue. Note that Hessian $H \in \mathbb{R}^{D\times D},\ D=11330$, thus $H$ around $\theta^*_{GD}$ approximately meets the ill-conditioning requirement in Proposition~{\ref{thm:aniso_benefit}}.
Figure~\ref{fig:sgdnoise}(Middle) shows the projection coefficient estimated by $\hat{a} = \frac{u_1^T \Sigma u_1 \Tr H}{\lambda_1 \Tr\Sigma}$ along the trajectory of SGD. The plot indicates that the projection coefficient is in a descent scale comparing to $D^{2d-1}$, thus satisfying the second condition in Proposition~\ref{thm:aniso_benefit}.
Therefore, Proposition~\ref{thm:aniso_benefit} ensures that SGD would escape from minima $\theta^*_{GD}$ faster than GLD in order of $\Ocal (D^{2d-1})$, as shown in Figure~\ref{fig:sgdnoise}(Right).
An interesting observation is that in the later stage of SGD optimization, $\Tr(H\Sigma)$ becomes significantly ($10^7$ times) smaller than in the beginning stage, implying that SGD has already converged to minima being almost impossible to escape from. This phenomenon demonstrates the reasonability to employ $\Tr(H\Sigma)$ as an empirical indicator for escaping efficiency.

\subsection{SVHN and CIFAR-10}
We also provide experiments on SVHN and CIFAR-10 datasets with VGG11 in Figure~\ref{fig:generalization_svhn_cifar} and Figure~\ref{fig:escape_svhn_cifar}.
For CIFAR-10 we use the original datasets while we only use $2,5000$ training examples for SVHN to compromise with the computational burden of gradient descent.
We choose VGG over ResNet since it achieves decent performance without using batch normalization, which causes extra affects on analyzing the noise of SGD.
We re-estimate the noise structure of GLD dynamic and GLD diag every $10$ iterations to ease the computational burden.
Also, we only run GD, GLD const, GLD dynamic, GLD diag and SGD since the computational costs of these dynamics are relatively acceptable to our hardware.

From Figure~\ref{fig:generalization_svhn_cifar} we can see the generalization gap between SGD and other dynamics, which demonstrates that the magnitude of SGD cannot fully explain the performance of SGD.
Figure~\ref{fig:escape_svhn_cifar} shows the escaping behavior of SGD and other dynamics, the results are consistent with experiments on FashionMNIST.

\section{Discussions}
\label{sec:dis}

\textbf{Benefits of considering covariance structure.} Previous works on SGD for deep learning typically ignores the covariance structure, as we have shown in this work, which has significant effects on its dynamics behaviors and generalization performance as well. 
The key observation on connecting gradient noise structure with curvature of the loss landscape, especially near the minima, provides a new perspective for understanding why SGD can achieve good generalization in practice.
Our work is an initial attempt to reveal the non-negligible benefits of SGD's covariance structure. More theoretical explorations are needed along this direction.

\textbf{Effects of learning rate and batch size.}
As seen from the SGD dynamics in Eq.~(\ref{eq:sgd}), when the learning rate is too small or batch size is overly large, the magnitude of gradient noise will become small, and thus effects of covariance structure is not obvious as before. In these cases, SGD often needs long time for diffusion towards flat minima to obtain better solutions, as shown in existing research~\citep{keskar2016large,hoffer2017,jastrzkebski2017three}.

\textbf{Designing optimizers that help to generalize better.}
The derived indicator also sheds some light on designing the optimizers that might generalize better than SGD by adding the noise along the direction of the maximum eigenvector of Hessian. We leave the exploration regarding  this as future work.

\section{Conclusion}
\label{sec:con}

We theoretically investigate a general optimization dynamics with unbiased noise, which unifies various existing optimization methods, including SGD.
We provide some novel results on the behaviors of escaping from minima and its regularization effects.
A novel indicator is derived for characterizing the escaping efficiency.
Based on this indicator, two conditions are constructed for showing what type of noise structure is superior to isotropic noise in term of escaping.
We then analyze the noise structure of SGD in neural networks and find that it indeed satisfies the two conditions, thus explaining the widely known observation that SGD can escape from sharp minima efficiently toward flat ones that generalize well.
Various experimental evidence supports our arguments on the behavior of SGD and its effects on generalization. 
Our study also shows that isotropic noise helps little for escaping from sharp minima, due to the highly anisotropic nature of landscape.
This indicates that it is not sufficient to analyze SGD by treating it as an isotropic diffusion over landscape \citep{zhang2017hitting,mou2017generalization}.
A better understanding of this out-of-equilibrium behavior \citep{chaudhari2017stochastic} is on demand. 
\section*{Acknowledgement}
This  work  is  supported  by  National  Natural  Science  Foundation  of  China  (No.61806009),  Beijing Natural  Science  Foundation  (No.4184090), Beijing Academy of Artificial Intelligence (BAAI) and Intelligent  Manufacturing  Action  Plan  of  Industrial  Solid Foundation  Program  (No.JCKY2018204C004).

\bibliography{bibliography.bib}
\bibliographystyle{icml2019}

\appendix
\section{Derivations and Proofs for Main Paper}

\subsection{Derivation of Eq.~(11) in main paper}
\begin{proof}
The "mild smoothness assumptions" refers that $L(\theta_t) \in C^2$.
Then the Ito's lemma holds~\citep{oksendal2003stochastic}. Thus,
\begin{align}
    &\dif L(\theta_t) \\
    =& \left(-\down L^T \down L + \half \Tr \left(\Sigma_t^{\half}H_t\Sigma_t^{\half}\right) \right) \dif t
    + \down L^T \Sigma_t^{\half} \dif W_t \\
    =& \left(-\down L^T \down L + \half \Tr \left(H_t\Sigma_t\right) \right) \dif t
    + \down L^T \Sigma_t^{\half} \dif W_t.
\end{align}
Taking expectation with respect to the distribution of $\theta_t$, we have
\begin{equation}
    \dif \Ebb_{\theta_t} L(\theta_t) = \Ebb \left( -\down L^T \down L + \half \Tr (H_t \Sigma_t)\right) \dif t,
\end{equation}
since the expectation of Brownian motion is zero.

Thus the solution of $\Ebb_{\theta_t} L(\theta_t)$ is,
\begin{equation}
    \Ebb L(\theta_t) = L(\theta_0) - \int_{0}^{t} \Ebb \left(\down L^T \down L\right)
    + \int_{0}^{t} \half \Ebb \Tr (H_t\Sigma_t) \dif t.
\end{equation}
\end{proof}

\subsection{Derivation of Eq.~(13) in main paper}
\begin{proof}
Without loss of generality, we assume that $L(\theta_0) = 0$.

For multivariate Ornstein-Uhlenbeck process, when $\theta_0 = 0$ is an constant, $\theta_t$ follows a multivariate Gaussian distribution~\citep{oksendal2003stochastic}.

For symmetric matrix $A$, let
\begin{equation}
    e^{A} := U^T \text{diag}(e^{\lambda_1}, \dots, e^{\lambda_n}) U,
\end{equation}
where $\lambda_1, \dots, \lambda_n$ and $U$ are the eigenvalues and eigenvector matrix of $A$.

Consider change of variables $\theta \to \phi(\theta, t) = e^{Ht}\theta_t$.
Note that,
\begin{equation}
    \od{e^{Ht}}{t} = H e^{Ht}.
\end{equation}
Thus by applying Ito's lemma, we have
\begin{equation}
    \dif \phi(\theta_t, t) = e^{Ht}\Sigma^{\half} \dif W_t,
\end{equation}
which we can integrate form $0$ to $t$ to obtain
\begin{equation}
    \theta_t = 0 + \int_0^t e^{H(s-t)}\Sigma^{\half}\dif W_s.
\end{equation}

The expectation of $\theta_t$ is zero. And by Ito's isometry~\citep{oksendal2003stochastic}, the covariance of $\theta_t$ is,
\begin{align}
    &\Ebb \theta_t \theta_t^T \\
    =& \Ebb \left[\int_0^t e^{H(s-t)}\Sigma^{\half} \dif W_s \left(\int_0^t e^{H(r-t)}\Sigma^{\half} \dif W_r\right)^T\right] \\
    =& \Ebb \left[\int_0^t e^{H(s-t)}\Sigma^{\half}\Sigma^{\half}e^{H(s-t)}\dif s \right] \\
    =& \Ebb \left[\int_0^t e^{H(s-t)}\Sigma e^{H(s-t)}\dif s \right] \\
    =& \int_0^t e^{H(s-t)}\Sigma e^{H(s-t)}\dif s. 
\end{align}
The last equation is because $H$ and $\Sigma$ are both constant.

Therefore
\begin{align}
\Ebb L(\theta_t) &= \half \Ebb \Tr \left(\theta_t^T H \theta_t\right) \\
                 &= \half \Tr \left(H \Ebb\theta_t \theta_t^T\right) \\
                 &= \half \int_0^t \Tr \left(H e^{H(s-t)} \Sigma e^{H(s-t)}\right) \dif s \\
                 &= \half \int_0^t \Tr \left(e^{H(s-t)} H\Sigma e^{H(s-t)}\right) \dif s \label{eq:commute} \\
                 &= \half \int_0^t \Tr \left(e^{2H(s-t)} H\Sigma\right) \dif s \\
                 &= \half \Tr \left(\half H^{-1} \left(I-e^{-2Ht}\right) H\Sigma\right) \\
                 &= \frac{1}{4} \Tr \left( \left(I-e^{-2Ht}\right)\Sigma\right).
\end{align}
Eq.~\eqref{eq:commute} holds since $H$ is symmetric.
Further, by Taylor's expansion we have
\begin{equation}
    \Ebb L(\theta_t) = \frac{1}{4} \Tr \left( \left(I-e^{-2Ht}\right)\Sigma\right) =\frac{t}{2}\Tr(H\Sigma).
\end{equation}
\end{proof}

\subsection{Proof of Proposition~1}
\begin{proof}
$\Tr(H\Sigma)$ can be decomposed as
\begin{equation}
\Tr(H\Sigma) = \sum_{i=1}^D \lambda_i u_i^T \Sigma u_i.
\end{equation}

Thus by the conditions of Proposition 1, we can bound $\Tr(H\Sigma)$ as
\begin{equation}
    \Tr(H\Sigma) \ge u_1^T \Sigma u_1\ge a\lambda_1\frac{\Tr \Sigma}{\Tr H}.
\end{equation}
On the other hand,
\begin{equation}
    \Tr(H\bar{\Sigma}) = \frac{\Tr\Sigma}{D}\Tr H.
\end{equation}
Thus,
\begin{equation}
\begin{aligned}
    \frac{\Tr(H\Sigma)}{\Tr(H\bar{\Sigma})} &\ge \frac{a \lambda_1 D}{\left(\Tr H\right)^2} \ge \frac{a \lambda_1 D}{\left(k\lambda_1 + (D-k)D^{-d}\lambda_1\right)^2} \\
    &= \Ocal\left(aD^{2d-1}\right).
\end{aligned}
\end{equation}
\end{proof}

\subsection{Proof of Proposition~2 in main paper}
\begin{proof}
For simplicity, we define 
\begin{equation}
    \bar{f}(x;\theta) := \phi\circ f(x;\theta) \in [\delta, 1-\delta],
\end{equation}
and
\begin{equation}
    \ell (x,y; \theta) = \half (\bar{f}(x;\theta) - y)^2.    
\end{equation}
Then the loss function becomes $L(\theta) = \Ebb_{(x,y)} \ell(x,y;\theta)$.

Since both $f$ and $\phi$ are piecewise linear, $\bar{f}(x;\theta)$ is also piece-wise linear with respect to $\theta$.
Thus the Hessian of $\bar{f}$ is zero almost everywhere.

We calculate the gradient and the Hessian of the loss:
\begin{equation}
    \down_{\theta} L(\theta) = \Ebb (\bar{f}(x;\theta) - y) \down_{\theta} \bar{f}(x;\theta);
\end{equation}
\begin{align}
    H(\theta) &= \down^2_{\theta} L(\theta) \\
    &= \Ebb \down_{\theta} \bar{f} (x;\theta) \cdot \down_{\theta} \bar{f} (x;\theta)^T + \Ebb (\bar{f}(x;\theta)-y)\down^2_{\theta} \bar{f}(x;\theta) \\
    &= \Ebb \down_{\theta} \bar{f} (x;\theta) \cdot \down_{\theta} \bar{f} (x;\theta)^T. \quad \text{almost everywhere.}
\end{align}
The last equation holds almost everywhere, since $\bar{f}(x;\theta)$ is piece-wise linear and its Hessian is zero almost everywhere. 

On the other hand, the Fisher is
\begin{align}
    F(\theta) &= \Ebb \down_{\theta} \ell(x,y;\theta) \cdot \down_{\theta} \ell(x,y;\theta)^T \\
    &= \Ebb (\bar{f}(x;\theta) - y)^2 \down_{\theta} \bar{f}(x;\theta) \cdot \down_{\theta} \bar{f}(x;\theta)^2.
\end{align}

(1)
Note that $\bar{f} \in [\delta, 1-\delta]$ and $y\in \{0,1\}$, thus
\begin{equation}
    (\bar{f}(x;\theta) - y)^2 \ge \delta^2. 
\end{equation}
Therefore
\begin{equation}
    F(\theta) \succeq \Ebb \delta^2 \down_{\theta} \bar{f}(x;\theta) \cdot \down_{\theta} \bar{f}(x;\theta)^2 = \delta^2 H(\theta),
\end{equation}
holds almost everywhere.

(2)
Around the minima where $\theta \in \{\theta : \norm{f(x;\theta)-y} \le \delta + \epsilon, \forall (x,y) \}$, we have
\begin{equation}
    (\bar{f}(x;\theta) - y)^2 \le (\delta+\epsilon)^2. 
\end{equation}
Therefore
\begin{equation}
    F(\theta) \preceq \Ebb (\delta+\epsilon)^2 \down_{\theta} \bar{f}(x;\theta) \cdot \down_{\theta} \bar{f}(x;\theta)^2 = (\delta+\epsilon)^2 H(\theta),
\end{equation}
holds almost everywhere around the minima.
\end{proof}

\subsection{Proof of Proposition~3 in main paper}
\begin{proof}
We only consider $\theta$ around the minima $\theta^*$ such that $\{\theta : \norm{\phi \circ f(x;\theta)-y} \le \delta + \epsilon, \forall (x,y) \}$.
On the other hand by construction $\norm{\phi \circ f(x;\theta)-y} \ge \delta$.
Thus according to Proposition~2,
\begin{equation}
\delta^2 H(\theta) \preceq F(\theta) \preceq (\delta+\epsilon)^2 H(\theta)
\label{eq:h_f}
\end{equation}
holds almost everywhere.

Thus let $\lambda(\theta)$ and $u(\theta)$ being the maximal eigenvalue and its corresponding eigenvector of $H(\theta)$,
\begin{equation}
    u(\theta)^T F(\theta) u(\theta) \ge \delta^2 u(\theta)^T H(\theta) u(\theta) = \delta^2 \lambda(\theta).
\end{equation}
Since at the minimal $\theta^*$ the Hessian is not zero, thus there is a positive value $\lambda^* >0 $ such that $\lambda(\theta^*) > \lambda^* > 0$.
Therefore by the continuity of $H(\theta)$, there are $\epsilon_1, \delta_1$, such that,
\begin{equation}
    \lambda(\theta) > \lambda^* - \epsilon_1 > 0, \quad \forall \norm{\theta-\theta^*} \le \delta_1.
\end{equation}

By Taylor's expansion,
\begin{equation}
\begin{aligned}
    \down L(\theta) &= \down L(\theta^*) + H(\theta^*)(\theta-\theta^*)+o(\theta-\theta^*) \\
    &= H(\theta^*)(\theta-\theta^*)+o(\theta-\theta^*).
\end{aligned}
\end{equation}
Hence,
\begin{equation}
    \norm{\down L(\theta)}_2^2 \le \norm{H(\theta^*)}_2^2 \norm{\theta-\theta^*}_2^2 + o\left(\norm{\theta-\theta^*}_2^2\right).
\end{equation}
Therefore, for all $\theta$ such that
\begin{align}
\norm{\theta-\theta^*}_2 &\le \frac{\sqrt{\delta^2\delta_2 (\lambda^* - \epsilon_1)}}{\norm{H(\theta^*)}_2} \\
&\le \frac{\sqrt{\delta^2\delta_2 \lambda(\theta)}}{\norm{H(\theta^*)}_2} \\
&\le \frac{\sqrt{\delta_2 u(\theta)^T F(\theta) u(\theta)}}{\norm{H(\theta^*)}_2},
\end{align}
we have
\begin{equation}
    \norm{\down L(\theta)}_2^2 \le \delta_2 u(\theta)^T F (\theta) u(\theta) + o\left(\abs{\delta_2 u(\theta)^T F(\theta) u(\theta)}\right).
\end{equation}

On the other hand, by definition, the gradient covariance $\Sigma$ and Fisher $F$ has the following relationship,
\begin{equation}
\begin{aligned}
    \Sigma (\theta) &= \Ebb (\down\ell(x,y;\theta) - \down L(\theta))\cdot (\down\ell(x,y;\theta) - \down L(\theta))^T \\
    &= \Ebb \down\ell(x,y;\theta)\cdot \down\ell(x,y;\theta)^T - \down L(\theta)\down L(\theta)^T \\
    &= F(\theta) - \down L(\theta)\down L(\theta)^T.
\end{aligned}
\end{equation}

Thus,
\begin{align}
    &\frac{u(\theta)^T \Sigma(\theta) u(\theta)}{\Tr\Sigma(\theta)} \\
    =& \frac{u(\theta)^T F(\theta) u(\theta) - u(\theta)^T \down L(\theta) \down L(\theta)^T u(\theta)}{\Tr F(\theta)-\Tr(\down L(\theta) \down L(\theta)^T)} \\
    =& \frac{u(\theta)^T F(\theta) u(\theta) - \norm{\down L(\theta)}_2^2}{\Tr F(\theta) - \norm{\down L(\theta)}_2^2} \\
    \ge& \frac{u(\theta)^T F(\theta) u(\theta) - \norm{\down L(\theta)}_2^2}{\Tr F(\theta)} \\
    =& \frac{u(\theta)^T F(\theta) u(\theta)}{\Tr F(\theta)} \left(1-\frac{\norm{\down L(\theta)}_2^2}{u(\theta)^T F(\theta) u(\theta)}\right) \\
    \ge& \frac{u(\theta)^T F(\theta) u(\theta)}{\Tr F(\theta)} \left(1-\delta_2-o\left(\abs{\delta_2}\right)\right) \\
    \ge& \frac{u(\theta)^T F(\theta) u(\theta)}{\Tr F(\theta)} \left(1-2\delta_2\right).
\end{align}

Note that Eq.~\eqref{eq:h_f} indicates that
\begin{align}
    & \forall u,\quad u^T (F(\theta) - \delta^{2} H(\theta)) u \ge 0 \\
    & \text{and }\quad \Tr((\delta+\epsilon)^{2} H(\theta) - F(\theta)) \ge 0.
\end{align}
Thus
\begin{align}
\frac{u(\theta)^T F(\theta) u(\theta)}{\Tr F(\theta)} &\ge \frac{\delta^2  u(\theta)^T H(\theta) u(\theta)}{(\delta+\epsilon)^2 \Tr H(\theta)} \\
& = \frac{\delta^2  \lambda(\theta)}{(\delta+\epsilon)^2 \Tr H(\theta)}.
\end{align}

Therefore for all $\theta$ in the set of
\begin{equation}
\begin{aligned}
&\left\{ \norm{\phi \circ f(x;\theta)-y} \le \delta + \epsilon, \forall (x,y)\right\} \\
\cap &\left\{\norm{\theta - \theta^*} \le \delta^1 \right\} \\
\cap &\left\{\norm{\theta-\theta^*}_2 \le \frac{\sqrt{\delta^2\delta_2 (\lambda^* - \epsilon_1)}}{\norm{H(\theta^*)}_2} \right\},
\end{aligned}
\end{equation}
we have
\begin{align}
    \frac{u(\theta)^T \Sigma(\theta) u(\theta)}{\Tr \Sigma(\theta)} &\ge \frac{u(\theta)^T F(\theta) u(\theta)}{\Tr F(\theta)} (1-2\delta_2)\\
    &\ge \frac{(1-2\delta_2)\delta^2}{(\delta+\epsilon)^2} \frac{\lambda(\theta)}{\Tr H(\theta)}.
\end{align}

\end{proof}

\section{About the non-convexity of the model in Proposition~2 in main paper}
Suppose we only have one training data $\{x=(1,1); y=1\}$, and the threshold activation is
\begin{equation}
    \phi(f) = \min\{\max\{f, 0.1\}, 0.9\}.
\end{equation}
Thus the loss is
\begin{equation}
    L(w_1, w_2) = (\phi(relu(w_1) - relu(w_2)) - 1)^2.
\end{equation}
Hence
\begin{equation}
    \begin{aligned}
    & L(1, 0) = 0.01 \\
    & L(0, 1) = 0.81 \\
    & L(0.5, 0.5) = 0.81.
    \end{aligned}
\end{equation}
Therefore
\begin{equation}
    \half L(1,0) + \half L(0,1) < L(0.5, 0.5),
\end{equation}
which means that $L$ is not convex.

It is also easy to see that $L$ has multiple minima.

\section{Additional experiments}
\subsection{Dominance of noise over gradient}
\begin{figure}
    \centering
    \includegraphics[width=0.8\columnwidth]{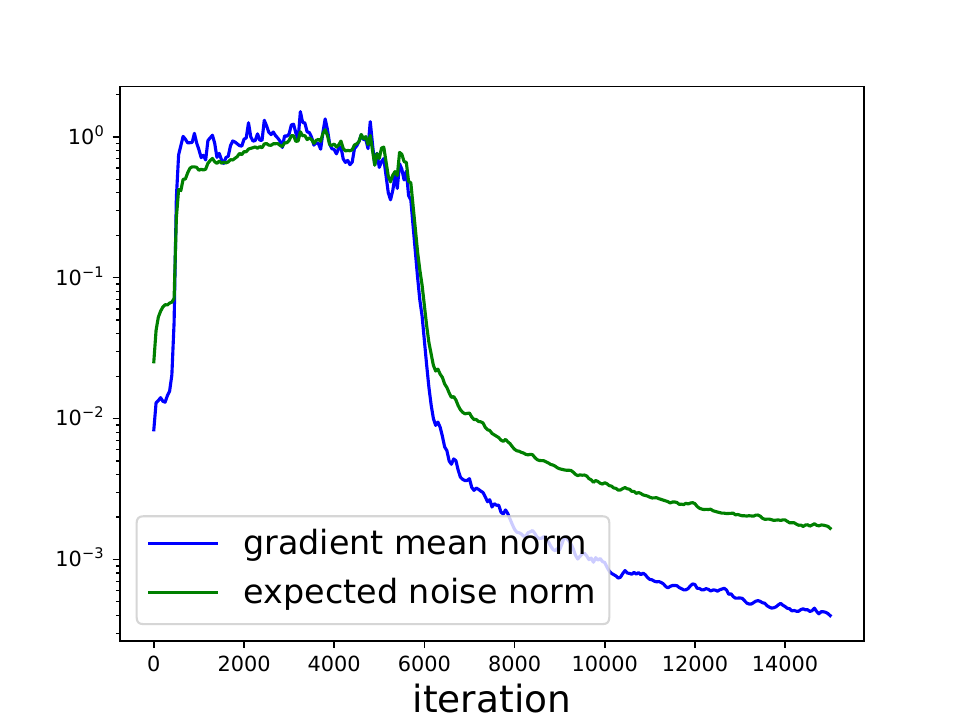}
    \caption{$L_2$ norm of gradient mean vs. the expected norm of noise during the training using SGD.
    The dataset and model are same as the experiments of FashionMNIST in main paper, or as in Section~\ref{sec:mnistsetup}}
    \label{fig:mnist_noisenorm}
\end{figure}

Figure~\ref{fig:mnist_noisenorm} shows the comparison of gradient mean and the expected norm of noise during training using SGD.
The dataset and model are same as the experiments of FashionMNIST in main paper, or as in Section~\ref{sec:mnistsetup}.
From Figure~\ref{fig:mnist_noisenorm}, we see that in the later stage of SGD optimization, the magnitude of noise indeed dominates that of gradient.

These experiments are implemented by TensorFlow $1.5.0$.

\subsection{The first $50$ iterations of FashionMNIST experiments in main paper}
\begin{figure}
\centering
\begin{tabular}{cc}
    \includegraphics[width=0.42\columnwidth]{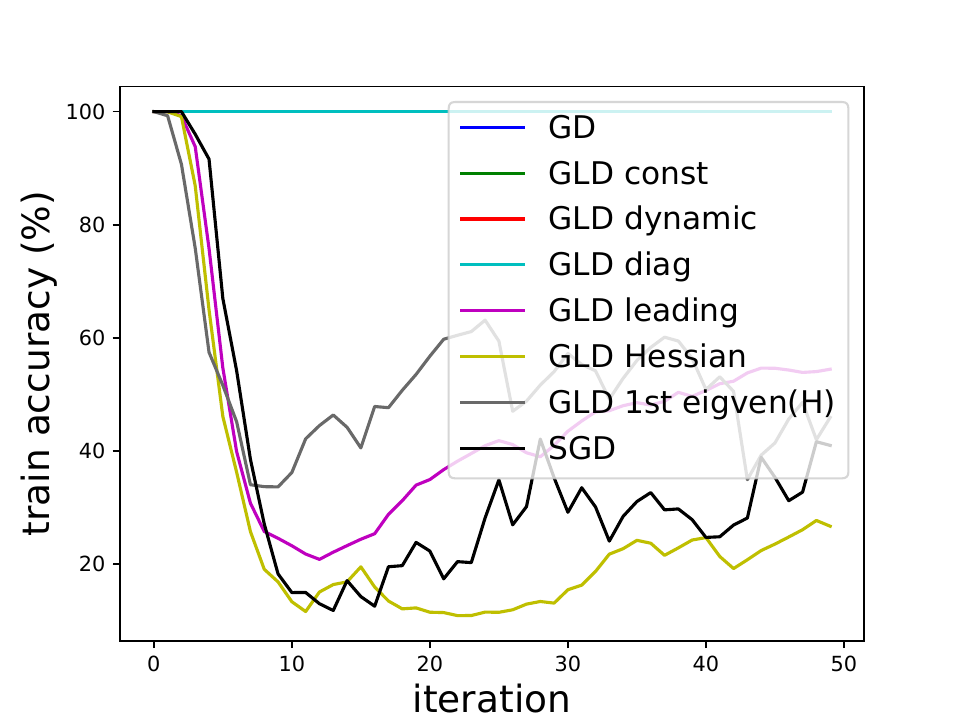} &  \includegraphics[width=0.42\columnwidth]{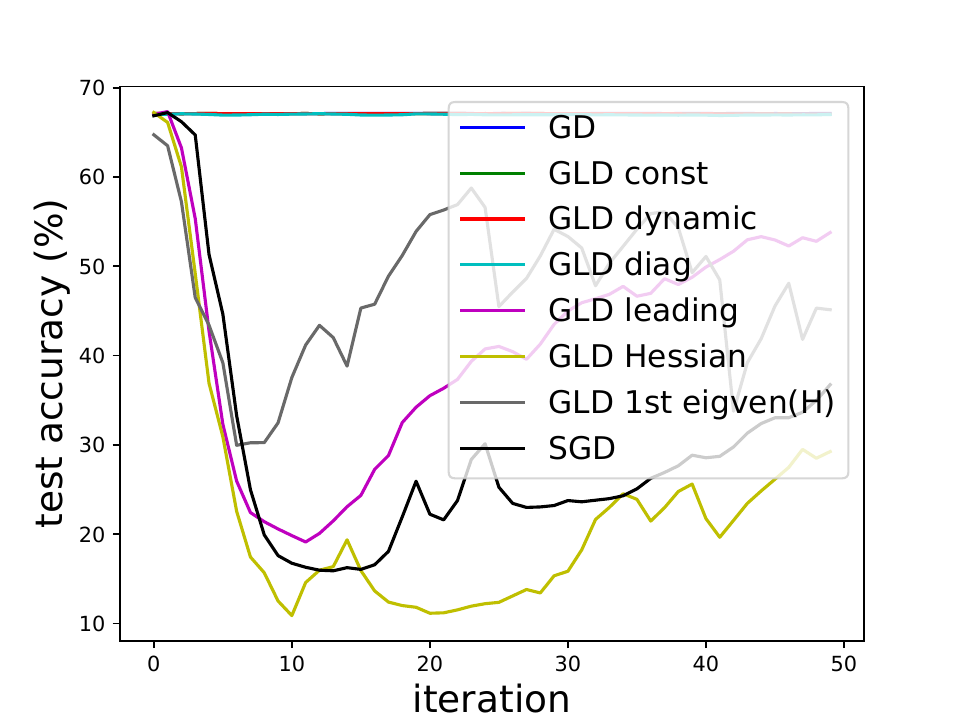}
\end{tabular}
\caption{\small The fisrt $50$ iterations of FashionMNIST experiments in main paper.
Compared dynamics are initialized at $\theta^*_{GD}$ found by GD.
The learning rate is same for all the compared methods, $\eta_t = 0.07$, and batch size $m=20$.
\textbf{Left}: Training accuracy versus iteration.
\textbf{Right}: Test accuracy versus iteration.}
\label{fig:mnist_50}
\end{figure}

Figure~\ref{fig:mnist_50} shows the first $50$ iterations of FashionMNIST experiments in main paper.
We observe that SGD, GLD 1st eigvec($H$), GLD Hessian and GLD leading successfully escape from the sharp minima found by GD, while GLD diag, GLD dynamic, GLD const and GD do not.

These experiments are implemented by TensorFlow $1.5.0$.

\section{Detailed setups for experiments in main paper}

\subsection{Two-dimensional toy example}

\begin{figure}
    \begin{center}
    \includegraphics[width=0.8\columnwidth]{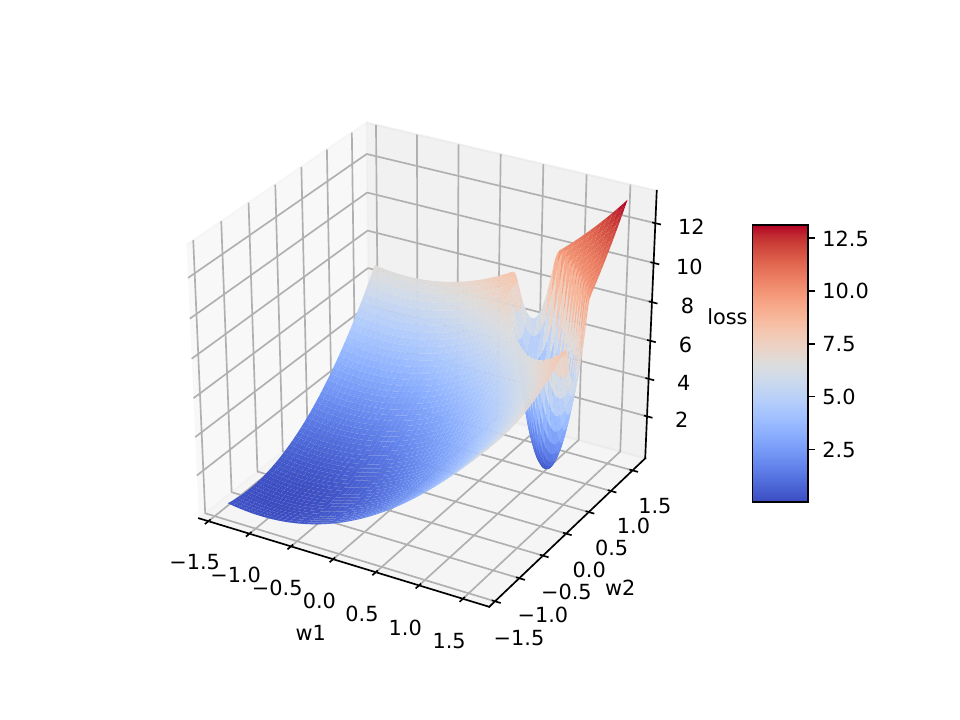}
    \caption{Constructed 2-dimensional surface in main paper.}
    \label{fig:surface}
    \end{center}
\end{figure}

\paragraph{Loss Surface}
The loss surface $L(w_1, w_2)$ is constructed by,
\begin{align*}
    &s_1 = w_1 - 1 - x_1, \\
    &s_2 = w_2 - 1 - x_2, \\
    &\ell(w_1, w_2; x_1, x_2) = \min\{10(s_1 \cos \theta  - s_2 \sin \theta )^2 \\
    &+ 100 (s_1 \cos \theta  + s_2 \sin \theta )^2,
       (w_1-x_1+1)^2 + (w_2-x_2+1)^2\}, \\
    &L(w_1, w_2) = \frac{1}{N}\sum^{N}_{k=1} \ell(w_1, w_2; x^k_1, x^k_2),
\end{align*}
where
\begin{align*}
    \theta &= \frac{1}{4}\pi,\\
    N &= 100,\\
    x^k &\sim \Ncal (0, \Sigma),\quad \Sigma = \begin{pmatrix}
        \cos \theta & \sin\theta \\
        -\sin\theta & \cos\theta
    \end{pmatrix}.
\end{align*}
Note that $\Sigma$ is the inverse of the Hessian of the quadric form generalizeing the sharp minima.
And the 3-dimensional plot of the loss surface is shown in Figure~\ref{fig:surface}.

\paragraph{Hyperparameters}
All learning rates are equal to $0.005$. All dynamics concerned are tuned to share the same expected square norm, $0.01$. The number of iteration during one run is $500$.

These experiments are implemented by PyTorch $0.3.0$.

\subsection{One hidden layer network}
\paragraph{Hyperparameters}
The $\delta$ is set to be $0.001$.
The learning rate is $0.001$.
The optimizer is Adam for fast convergence, which does not affect our point on studying $\Tr (H\Sigma)$.

The code is implemented in TensorFlow 1.9.0.

\subsection{FashionMNIST with corrupted labels}
\label{sec:mnistsetup}
\paragraph{Dataset}
Our training set consists of $1,200$ examples randomly sampled from original FashionMNIST training set, and we further specify $200$ of them with randomly wrong labels. The test set is same as the original FashionMNIST test set.

\paragraph{Model}
Network architecture:
\begin{align*}
    \text{input} &\Rightarrow \text{conv1} \Rightarrow \text{max\_pool} \Rightarrow \text{ReLU} \Rightarrow \text{conv2} \\
    &\Rightarrow \text{max\_pool} \Rightarrow \text{ReLU}
    \Rightarrow \text{fc1} \Rightarrow \text{ReLU} \\
    &\Rightarrow \text{fc2} \Rightarrow \text{output}.
\end{align*}
Both two convolutional layers use $5\times 5$ kernels with $10$ channels and no padding. The number of hidden units between fully connected layers are $50$. The total number of parameters of this network are $11,330$.

\paragraph{Training details}
\begin{itemize}
    \item \textbf{GD}: Learning rate $\eta=0.1$. We tuned the learning rate (in diffusion stage) in a wide range of $\{0.5, 0.2, 0.15, 0.1, 0.09, 0.08, \dots, 0.01\}$ and no improvement on generalization.
    \item \textbf{GLD constant}: Learning rate $\eta=0.07$, noise std $\sigma=10^{-3}$. We tuned the noise std in range of $\{10^{-1}, 10^{-2}, 10^{-3}, 10^{-4}, 10^{-5}\}$ and no improvement on generalization.
    \item \textbf{GLD dynamic}: Learning rate $\eta=0.07$.
    \item \textbf{GLD diagnoal}: Learning rate $\eta=0.07$.
    \item \textbf{GLD leading}: Learning rate $\eta=0.07$, number of leading eigenvalues $k=20$, batchsize $m=20$. We first randomly divide the training set into $60$ mini batches containing $20$ examples, and then use those minibatches to estimate covariance matrix.
    \item \textbf{GLD Hessian}: Learning rate $\eta=0.07$, number of leading eigenvalues $=20$, update frequence $f=10$. Do to the limit of computational resources, we only update Hessian matrix every $10$ iterations. But add Hessian generated noise every iteration. And to the same reason, we simplily set the coefficent of Hessian noise to $\sqrt{\Tr H / m\Tr \Sigma}$, to avoid extensively tuning of hyperparameter.
    \item \textbf{GLD 1st eigvec($H$)}: Learning rate $\eta = 0.07$, as for GLD Hessian, and we set the coefficient of noise to $\sqrt{\lambda_1 / m\Tr \Sigma}$, where $\lambda_1$ is the first eigenvalue of $H$.
    \item \textbf{SGD}: Learning rate $\eta=0.07$, batchsize $m=20$.
\end{itemize}

\paragraph{Estimation of Sharpness}
The sharpness are estimated by 
\begin{equation}
\frac{1}{M} \sum_{j=1}^M L(\theta+\nu_j) - L(\theta),\quad \nu_j \sim \mathcal{N}(0,\delta^2I),
\end{equation}
with $M=1,000$ and $\delta=0.01$.

These experiments are implemented by TensorFlow $1.5.0$.

\subsection{SVHN and CIFAR-10}
\paragraph{Dataset}
For SVHN experiments, we use $2,5000$ examples for training and $7,5000$ examples for test, to compromise with the computational burden of gradient descent.
And for CIFAR-10 experiments, we use standard CIFAR-10 datasets.
We do not use data augmentation since it could cause uncontrollable affects on analyzing SGD noise.

\paragraph{Model}
Standard VGG11 network without any regularizations including dropout, batch normalization, weight decay, etc.
The total number of parameters of this network is $9,750,922$.

We choose VGG11 instead of ResNet because VGG11 achieves good generalization performance without using \emph{Batch Normalization}, which has a subtle impact on SGD noise.

\paragraph{Training details}
Learning rates $\eta_t=0.05$ are fixed for all optimizers, which is tuned for the best generalization performance of GD.
The batch size of SGD is $m=100$.
The noise std of GLD constant is $\sigma=10^{-3}$, which is tuned to best.
Due to computational limitation, we only conduct experiments on GD, GLD const, GLD dynamic, GLD diag and SGD.

\paragraph{Estimation of Sharpness}
The sharpness are estimated by
\begin{equation}
\frac{1}{M} \sum_{j=1}^M L(\theta+\nu_j) - L(\theta),\quad \nu_j \sim \mathcal{N}(0,\delta^2I),
\end{equation}
with $M=100$ and $\delta=0.01$.

These experiments are implemented by PyTorch $1.0.0$.

\end{document}